\documentclass[authoryear,final,3p,times,review,preprint,onecolumn]{elsarticle}
\usepackage[labelfont=bf]{caption}
\usepackage{amssymb}
\usepackage{apalike}
\usepackage[longtable]{multirow}
\usepackage{longtable}
\usepackage{array}
\usepackage{colortbl}
\usepackage{amsmath}
\usepackage{graphicx}
\usepackage{booktabs}
\usepackage{ragged2e}
\usepackage[hidelinks, colorlinks=true]{hyperref}
\usepackage{tabularray}
\usepackage{float}
\usepackage{subcaption} 
\usepackage{apalike}
\begin{document}

\begin{frontmatter}



\title{AI in Supply Chain Risk Assessment: A Systematic Literature Review and Bibliometric Analysis}


%\author{Anonymous Author(s)}

\author[inst1]{Md Abrar Jahin\corref{cor1}}
\ead{jahin@usc.edu}
\author[inst2]{Saleh Akram Naife}
\ead{naife.iem.kuet@gmail.com}
\author[inst2]{Anik Kumar Saha}
\ead{anikkumarsaha1999@gmail.com}

\affiliation[inst1]{organization={Thomas Lord Department of Computer Science, Viterbi School of Engineering, University of Southern California},
    city={Los Angeles},
    state={CA},
    postcode={90089},
    country={USA}}

\affiliation[inst2]{organization={Department of Industrial Engineering and Management},Department and Organization
         addressline={Khulna University of Engineering \& Technology (KUET)}, 
         city={Khulna},
        postcode={9203}, 
         country={Bangladesh}}

\author[inst3]{and M. F. Mridha\corref{cor1}}
\ead{firoz.mridha@aiub.edu}

\affiliation[inst3]{organization={Department of Computer Science},Department and Organization
        addressline={American International University-Bangladesh (AIUB)}, 
        city={Dhaka},
        postcode={1229}, 
        country={Bangladesh}}
\cortext[cor1]{Corresponding author.}

\begin{abstract}
Supply chain risk assessment (SCRA) is pivotal for ensuring resilience in increasingly complex global supply networks. While existing reviews have explored traditional methodologies, they often neglect emerging artificial intelligence (AI) and machine learning (ML) applications and mostly lack combined systematic and bibliometric analyses. This study addresses these gaps by integrating a systematic literature review with bibliometric analysis, examining 1,903 articles (2015–2025), with 54 studies selected through PRISMA guidelines. Our findings reveal that ML models, including Random Forest, XGBoost, and hybrid approaches, significantly improve risk prediction accuracy and adaptability in post-pandemic contexts. The bibliometric analysis identifies key trends, influential authors, and institutional contributions, highlighting China and the United States as leading research hubs. Practical insights emphasize the integration of explainable AI (XAI) for transparent decision-making, real-time data utilization, and blockchain for traceability. The study underscores the necessity of dynamic strategies, interdisciplinary collaboration, and continuous model evaluation to address challenges such as data quality and interpretability. By synthesizing AI-driven methodologies with resilience frameworks, this review provides actionable guidance for optimizing supply chain risk management, fostering adaptability, and informing future research in evolving risk landscapes.
\end{abstract}


% Artificial Intelligence (AI); Supply chain risk assessment; Bibliometric analysis; Content analysis; Systematic literature review

\end{frontmatter}
\section{Introduction}
\label{sec:introduction}
Supply chain management (SCM) has long grappled with various challenges from various sources. Past outbreaks of infectious diseases, geological upheavals such as earthquakes, and other natural catastrophes have put supply chains (SCs) at risk, albeit to a limited extent \citep{govindan_decision_2020}. These incidents have illustrated how interrelationships within SCs have formed intricate risk contagion networks, making them susceptible to contagion effects \citep{agca_credit_2022}. The cascade effect, characterized by risk spillover from one enterprise to another, exacerbates the challenges, culminating in a chain reaction of SC Resilience (SCR) disasters \citep{roukny_interconnectedness_2018}. Across the globe, companies confront formidable challenges across all stages of their SCs. Suppliers failing to meet delivery obligations, unpredictable shifts in customer demands, and episodes of panic buying are just a few examples of the hurdles companies face \citep{ivanov_predicting_2020}. In pursuing market leadership, enterprises must also contend with the complexity introduced by the management and launch of innovation projects, further complicating SCM \citep{kwak_investigating_2018}. Furthermore, digital transformation has ushered in a wave of technological advancements in SC operations \citep{kwak_investigating_2018}.

However, the most significant disruption in recent times occurred during the COVID-19 pandemic, profoundly impacting global SCs. The disruptions encompassed not only the movement of people but also the flow of raw materials and finished goods, along with extensive disruptions in factory and SC operations \citep{sheng_covid-19_2021}. These disruptions ushered in unprecedented challenges for SC professionals as they grappled with an entirely new reality \citep{araz_data_2020, craighead_pandemics_2020}. The pandemic represents a distinctive form of SC disruption, distinct from natural or man-made disasters (e.g., the 2011 earthquake in Japan and the 9/11 attacks) and disruptions driven by evolving technologies and changing customer attitudes \citep{zhang_evolution_2020}. While existing guidance exists for predicting, managing, and responding to these disruptions, the challenges brought about by COVID-19 have underscored the paramount importance of risk management, unlike any previous disruption \citep{bode_understanding_2011,craighead_pandemics_2020}. Consequently, the concept of SCR in COVID-19 has emerged as a critical area necessitating further exploration and development.

SCs are not the sole victims in this precarious environment, as the ripple effect extends to upstream and downstream enterprises. Practical risk management programs are imperative for enterprises to avert SCR and improve SC risk management (SCRM) \citep{ivanov2021supply}. AI encompasses techniques that enable autonomous decision-making in complex, uncertain environments. As noted by \cite{baryannis_supply_2019}, an SCRM approach is considered artificially intelligent if it can autonomously decide on a course of action that leads to success in SCRM-related objectives while operating under a partially unknown SC environment. The AI techniques utilized in this context span from traditional symbolic AI, which relies on mathematical or knowledge-based problem representations, to sub-symbolic AI, including fuzzy systems and evolutionary computation, and statistical AI, which encompasses ML approaches. Among these, ML algorithms have showcased their ability to identify abnormal risk factors and derive predictive insights from historical data \citep{guo_integrated_2021, mohanty_financial_2021}. By leveraging ML, enterprises can detect risk factors, anticipate market demand, and anticipate potential risk scenarios \citep{punia_predictive_2020, wu2022credit}. ML's proficiency in handling non-linear relationships further bolsters its superiority over traditional linear models. Additionally, its aptitude for processing unstructured data, a challenge where traditional models falter, positions ML as a formidable tool for addressing time, cost, and resource constraints within SCs. ML has emerged as the most suitable technique for risk assessment in SC. Therefore, this review focuses specifically on ML-based approaches in the context of SC resilience assessment.

However, despite its immense potential, SC researchers have historically shown limited familiarity with ML compared to other SCM areas such as mathematical programming and stochastic optimization \citep{liu_two-period_2019, shahed_supply_2021}. Furthermore, the classification of ML algorithms remains unclear \citep{janiesch_machine_2021, xu_machine_2019}. Bridging this knowledge gap and exploring the value of ML for SCRM through interdisciplinary integration is a critical research need. Remarkably, there has been no prior effort to scrutinize the SCRM literature within an ML environment, in stark contrast to previous reviews centered on risk definition, classification, and management strategies. Companies should augment their analytical capabilities by utilizing organizational knowledge to improve SCR, thereby elevating their information capabilities \citep{wong_supply_2020}. As underscored by previous studies, AI-enabled technologies also promote innovations to improve SC performance (SCP) \citep{baryannis_supply_2019, nayal_are_2021}. The adaptation capabilities and information-processing prowess offered by AI techniques have the potential to improve SCP \citep{belhadi_manufacturing_2021}. Notably, AI has found application across diverse sectors, improving flexibility and communication and reducing undue fluctuations for successful project execution \citep{lalmi_conceptual_2021}. To navigate the impact of risks and disruptions, \citep{katsaliaki_supply_2022} advocates integrating three critical facets: long-term partnerships, IT applications for business improvement, and government policies that facilitate adaptability.

The contemporary SC landscape demands the integration of AI, specifically ML, to invigorate SCRA and SCRM. While traditional methods have merit, their limitations can be effectively addressed by AI, which offers predictive capabilities, nonlinear relationship analysis, and the processing of unstructured data. As SCs continue to evolve, AI-based risk assessment is imperative for fostering resilience and sustaining the efficiency and effectiveness of SC operations \citep{ganesh2022future}. Existing reviews of SCRA using ML often fail to offer a comprehensive view of evolving ML techniques, neglecting emerging trends and diverse applications, as well as non-journal sources. They also tend to overlook the role of ML in the response phases of SCRM and fail to adequately address data-related challenges \citep{robertson2025managing}.

In light of the limitations of traditional SCRA methods, there is growing interest in leveraging AI techniques to improve risk assessment practices. This paper presents a systematic literature review (SLR) focused on the application of AI in SCRA. The existing literature lacks a bibliometric analysis that incorporates recent and relevant articles on the AI-SCRA topic. Another limitation is that most of the literature contributed only to bibliometric or content analysis. We sought to close this gap by including both analyses in our study, using a systematic review methodology. The main objective of this review is to analyze the existing literature critically, identify research gaps, and provide insights into the use of AI techniques, such as ML, Deep Learning (DL), and natural language processing (NLP), to improve the accuracy and effectiveness of SCRA. Our paper has the following contributions-

\begin{enumerate}
    \item Our study provides a comprehensive review of AI and ML applications in SCRA, supported by bibliometric and co-citation analyses to capture recent advancements and trends in the field.
    \item We conduct content analysis to explore thematic trends, detailing the advantages, disadvantages, and specific application scenarios of various AI/ML models used in SCRA, offering insights into their suitability for different SC contexts.
    \item Our proposed AI-based framework guides practitioners in assessing SC risks, from data collection to model selection and interpretation, facilitating efficient risk management in dynamic SCs.
\end{enumerate}

The rest of the article is organized as follows: the ``\hyperref[sec:Literature Review]{Literature Review}" examines the current body of knowledge. The ``\hyperref[sec:methodology]{Methodology}" section details the study's design, including search strategies and selection criteria. The ``\hyperref[sec:biblio]{Bibliometric Analysis}" provides insights from bibliometric tools, with visualizations and observations on AI in SCRA. A structured framework for SCRA is given in ``\hyperref[sec:Framework for SCRA using AI]{Framework for SCRA using AI} section.  In the ``\hyperref[sec:AI Techniques]{Content Analysis of AI Techniques in SCRA}," the review discusses various AI techniques, highlighting their benefits, limitations, and real-world applications. Practical insights for the industry are provided in the ``\hyperref[sec:Managerial Implications]{Practical and Managerial Implications}" section, offering actionable guidance for practitioners. The review also outlines key ``\hyperref[sec:challenges]{Challenges and Limitations}" in this area, and suggests ``\hyperref[sec:Future Research Directions]{Future Research Directions}" for ongoing exploration. The paper concludes with a summary of findings and recommendations in the ``\hyperref[sec:conclusion]{Conclusion}."

\section{Literature Review}
\label{sec:Literature Review}
\cite{deiva_ganesh_supply_2022} conducted a comprehensive and descriptive literature study to identify AI and ML approaches in SCRM stages. This study examined SCRM research publications from three scientific databases from 2010 to 2021. They proposed a data analytics, simulation, and optimization framework that might produce a comprehensive SC risk identification, assessment, mitigation, and monitoring strategy. Hence, leveraging AI, Blockchain, and the Industrial Internet of Things (IIoT) to create smarter SCs can transform how firms handle uncertainty. This analysis of \cite{deiva_ganesh_supply_2022} does not evaluate blockchain, big data, and IIoT-related SC articles, which may limit information. Due to limited research, only recent English-language publications on SCRM and AI were included. \cite{nimmy_explainability_2022} did a thorough literature evaluation on operational risk assessment methods and whether AI explains them. The report suggested that risk managers use auditable SC operational risk management methodologies to understand why they should take a risk management action rather than just what to do. While they employed the SLR method, they could only examine some AI solutions for SCRM. They suggested hybridizing SC operational risk management and explainable AI (XAI) to obtain XAI-like characteristics in future research. \cite{li_developing_2023} analyzed COVID-19 SCR research history, present, and future. In particular, supervised ML classifies 1717 SCR papers into 11 subject categories. Each cluster was then studied in the context of COVID-19, indicating three related skills (interconnectedness, transformability, and sharing) on which enterprises could work to develop a more resilient SC post-COVID. Their data only came from Scopus' Core Collection, which might affect the results; thus, they suggested adding WoS and EBSCOhost to the evaluation \citep{xu_disruption_2020}. Given the rapid expansion of SCR research, they selected only English-language articles, which may have excluded valuable knowledge. They recommended network analysis to determine cluster linkages and SCR literary themes. 

\cite{naz_is_2022} examined the role of AI in building a robust and sustainable SC and offered optimal risk mitigation options. For review, 162 SCOPUS research publications were selected. Based on the nominated articles, Structural Topic Modeling identified various AI-related themes in SCR. AI research trends in SCR were examined using the R package for bibliometric analysis. They studied only journal articles, not conference papers, field reports, corporate reports, book chapters, etc. \cite{ni_systematic_2020} analyzed articles from 1998/01/01 to 2018/12/31 across five major databases to highlight the latest research trends in the field. ML applications in SCM were still developing due to a shortage of high-quality authors and low publication rates. 10 ML algorithms were extensively used in SCM, but their use was uneven across the most frequently reported SCM tasks. This paper has limitations, as it reviews only five popular databases and limits the articles for evaluation, potentially filtering some related articles. Second, only widely used ML methods in SCM were counted in this review; other ML techniques introduced to SCM may be helpful later. Low-frequency ML algorithms should be analyzed for further research in this area. Finally, their article included 32 well-established ML methods; however, some newly developed ML algorithms may be used in SCM after 2015.
\cite{baryannis_supply_2019} examined SC risk definitions and uncertainty. Then, a mapping analysis categorizes available literature by AI approaches and SCRM tasks. Most of the examined works focus on building and assessing mathematical models that account for various uncertainties and hazards, but less on establishing and analyzing the applicability of the proposed models. They found that only 9 of 276 studies (3\%) used comprehensive techniques that covered all three SCRM stages (identification, assessment, and response). \cite{yang_supply_2023} thoroughly examined the advancement of ML algorithms in SCRM by gathering 67 publications from 9 authoritative databases in the first half of 2021. They analyzed only English-language journal articles from 9 relevant academic databases, excluding conference papers, textbooks, and unpublished articles and notes. This analysis relied on keyword searches, which may have missed some work. \cite{schroeder_systematic_2021} conducted a comprehensive, multi-vocal literature review to provide a complete picture of our subject field. The SLR identified 533 papers in this area and examined 23 of them. The comprehensive literature review found that only a few examples of ML in SCRM have been reported in depth in the scientific literature, and these examples focus on manufacturing, transport, and the entire SC. \cite{shi_machine_2022} examined 76 works on credit risk utilizing statistical, ML, and DL methods over the last eight years. They offered a unique classification approach and performance rating for ML-driven credit risk algorithms utilizing public datasets. Data imbalance, dataset inconsistency, model transparency, and underuse of DL models are discussed. Their review found that most DL models outperform standard ML and statistical algorithms in credit risk prediction, and ensemble techniques outperform single models.

Previous studies in SCRM have shown limitations in evaluating certain technologies, such as blockchain, big data, and the IIoT, potentially limiting the scope of analysis. Our review addresses this gap by incorporating an analysis of these technologies within the SCRM context, providing a more comprehensive assessment. The literature recognizes a need to integrate SC operational risk management with XAI for future research. Our review contributes to closing this gap by exploring the integration of these approaches and discussing their implications for SCRM. Some studies have highlighted the importance of using multiple databases to comprehensively evaluate SCRM literature. Our review extends this by considering insights from the most reliable and prominent two databases, enhancing the breadth of our analysis. Previous research has found that only a small percentage of studies comprehensively cover all three stages of SCRM. Our review contributes to closing this gap by providing a holistic assessment of SCRM stages, including identification, assessment, and response strategies, leveraging insights from diverse studies.

\section{Methodology}
\label{sec:methodology}
\subsection{Study Design}
The research utilized a systematic approach to conduct a comprehensive literature review and a bibliometric analysis. This methodology aimed to scrutinize the intersection of AI/ML techniques and SCRA, providing insights into research trends, methodologies, and emerging themes. We formulated the following research questions to guide our SLR:

\begin{itemize}
    \item [RQ1:] What is the current state of research on applying AI/ML techniques in SCRM?
    \item [RQ2:] Which AI/ML techniques are commonly employed in SCRA, prediction, and mitigation? 
    \item [RQ3:] What are the key findings and trends identified in the literature regarding using AI/ML in SCRM? 
    \item [RQ4:] What are the research gaps and future directions in this field?
\end{itemize}

Our approach is guided by the structured methodology of \cite{denyer_producing_2009} for systematic literature reviews, ensuring rigor in the identification and synthesis of studies. Additionally, we adhered to the Preferred Reporting Items for Systematic Reviews and Meta-Analyses (PRISMA) framework to improve transparency and minimize bias in the study selection process. This approach aimed to identify and analyze research articles on integrating AI/ML techniques within SCRA.

\subsection{Search Strategy}
We conducted a comprehensive search using two major academic databases: Google Scholar and Web of Science (WoS) Core Collection. The WoS Core Collection, recognized as a high-quality digital literature database, has gained widespread acceptance among researchers and is currently regarded as one of the most appropriate databases for bibliometric analysis \citep{wang_education_2024}. Compared to other databases, WoS offers more extensive data coverage and provides data formats compatible with analysis tools such as VOSviewer. We used the ``Publish or Perish 8" \citep{harzing_publish_2016} software for the Google Scholar search to extract all data of the most relevant articles suitable for bibliometric analysis. The initial search on Google Scholar using \textit{Publish or Perish} yielded 1000 articles, and the WoS advanced search yielded 903 articles before specifying any date range. We employed the following search strings to retrieve relevant articles:
\begin{quote}
(ALL=( ``supply chain") AND (``risk*" OR ``credit risk*" OR ``risk assessment*" OR ``risk prediction" OR ``disruption*" OR ``disease outbreak*" OR ``post COVID resilienc*") AND (``artificial intelligence" OR ``machine learning" OR ``neural network*" OR ``deep learning" OR ``reinforcement learning" OR ``SVM" OR ``support vector machine" OR ``boosting" OR ``ensemble" OR ``bayesian network model" OR ``random forest" OR ``LSTM" OR ``long short-term memory"))) NOT (DT==(``PROCEEDINGS PAPER" OR ``BOOK CHAPTER" OR ``EDITORIAL MATERIAL" OR ``LETTER"))
\end{quote}

\begin{table}[!ht]
\centering
\caption{Inclusion and exclusion criteria and justifications}
\label{table:criteria}
\resizebox{\textwidth}{!}{%
\begin{tabular}{p{8cm}p{3cm}p{9cm}}
    \hline
    \textbf{Criterion} & \textbf{Inclusion/Exclusion} & \textbf{Justification} \\
    \hline
    Date range & 2015–2025 & Focuses on recent studies, ensuring relevance to current advancements in AI for SCRA. \\
    
    Language & English only & Ensures consistency in analysis and accessibility for an English-speaking academic audience. \\
    
    Duplicate removal & Exclusion & Avoids redundancy and ensures each study contributes uniquely to the analysis. \\
    
    Publisher reliability & Exclusion & Maintains academic quality by including only studies from reputable, peer-reviewed journals. \\
    
    Publication type & Exclusion of Q3/Q4, proceedings, book chapters, etc. & Focuses on high-impact research, prioritizing quality over quantity. \\
    
    Checking for relevance (title, abstract, and conclusion) & Exclusion & Ensures alignment with the study’s objectives based on content relevant to SC risk and AI. \\

    Unrelated methodologies & Exclusion & Removes studies that employ computational models or frameworks that do not align with the research scope. \\

    Limited generalizability & Exclusion & Excludes studies focusing on highly specific contexts that cannot be applied to broader AI-driven SCRA applications. \\
    \hline
\end{tabular}%
}
\end{table}

\subsection{Article Selection Process}
To ensure the selection of high-quality and relevant studies, we followed a systematic, multi-stage screening process guided by the PRISMA framework. The selection process was designed to minimize potential bias and ensure transparency at each stage. The stages are detailed as follows:

\subsubsection{Date Range Filtering:} We restricted our review to studies published between 2015 and 2025, excluding 23 articles, ensuring the inclusion of recent and relevant research. This step ensures that each study contributes uniquely to the analysis and aligns with the latest advancements in the field.

\subsubsection{Removal of Duplicates:} To avoid redundant analysis, we first identified and removed duplicate articles between the two databases, excluding 451 duplicate entries. This step is standard in systematic reviews to ensure each study contributes uniquely to the analysis.

\subsubsection{Filtering by Publisher Reliability:} We excluded 398 articles published by sources identified as less reliable, focusing only on those published in reputable publishers (IEEE, Springer Nature, Elsevier, Taylor \& Francis, MDPI, Sage, Wiley, Nature Portfolio, ACM, etc.). This criterion helps maintain high academic integrity and ensures the final selection is based on studies that meet established research quality standards.

\subsubsection{Non-English Exclusion:} Since this review focuses on an English-speaking academic audience, we excluded 3 articles not written in English. This ensures that the included studies are accessible and can be consistently interpreted without translation bias.

\subsubsection{Filtering by Publication Type:} To prioritize peer-reviewed journal articles, we excluded papers classified as Q3 or Q4 and proceedings papers, book chapters, editorials, and letters, removing 244 articles. This criterion ensures that our review is based on high-impact research published in top-quality journals, as lower-ranked journals or other types of publications may not always meet rigorous peer-review standards.

\subsubsection{Checking for Relevance (Title, Abstract, and Conclusion Screening):} The remaining articles were reviewed based on their title, abstract, and conclusions to assess their relevance to the research questions of this study. Articles that did not focus specifically on AI applications in SCRA were excluded. This step eliminated 515 articles.

\subsubsection{Exclusion of Studies with Unrelated Methodologies:} After relevance screening, we further analyzed the methodologies used in the remaining studies. Articles that employed computational models or frameworks outside our scope were removed. This step excluded 128 articles, ensuring the final selection focused on studies directly relevant to the research objectives.  

\subsubsection{Exclusion of Studies with Limited Generalizability:} We excluded 81 highly context-specific studies that lacked generalizability to broader AI-driven SCRA applications, leaving a final sample of 54 articles for in-depth review. These studies primarily focused on niche use cases that do not translate well to wider SCRA applications.

Each criterion was carefully selected to ensure that only high-relevance, high-quality studies were included, thereby minimizing potential bias. Using dual databases (Google Scholar and WoS) ensured comprehensive coverage, while the sequential exclusion criteria systematically filtered out irrelevant or low-quality studies. To demonstrate further transparency, Table \ref{table:criteria} outlines the inclusion and exclusion criteria and their justifications.

\subsection{Bias Mitigation}
To address concerns about potential bias, we followed a systematic approach that minimizes subjective selection at each stage:
\begin{enumerate}
\item By using two widely recognized databases, Google Scholar and WoS, we ensured broad initial coverage of the literature on AI in SCRA, reducing the likelihood of selection bias by capturing a diverse range of studies.

\item Each exclusion criterion was pre-defined and systematically applied, following a step-by-step approach that minimized subjective judgment. This ensures that all articles are assessed based on consistent standards.

\item The PRISMA framework was rigorously applied to eliminate irrelevant studies at multiple stages, further reducing the risk of bias by ensuring that only studies meeting high-quality and relevance criteria were retained.
\end{enumerate}

\subsection{Final Selection}
\begin{figure*}[!ht]
    \centering
    \includegraphics[width=1\linewidth]{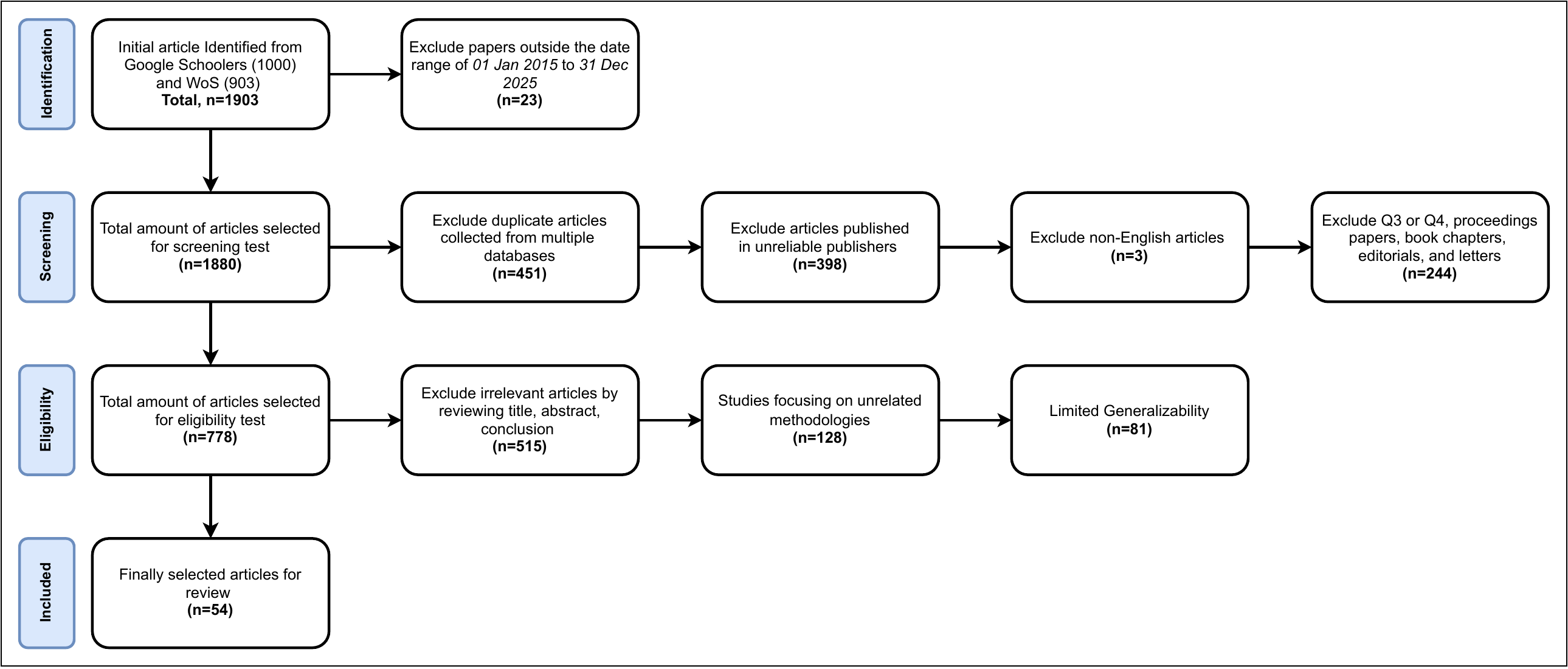}
    \caption{Prisma flow diagram for systematic literature review.}
    \label{fig:prisma_flow}
\end{figure*}

After applying all criteria, 54 articles were deemed relevant and of high quality for inclusion in this review. The selected articles cover various AI methodologies applied to different SC risk scenarios, providing a comprehensive foundation for our analysis. According to Figure \ref{fig:prisma_flow}, 778 articles were included in the bibliometric analysis to provide a broad understanding of the field, considering various quantitative metrics. The 54 shortlisted articles represent a more focused subset. We selected them based on additional qualitative criteria, such as relevance, methodology, and contribution to the research.

\subsection{Data Extraction and Synthesis}
The selected articles underwent detailed analysis, focusing on methodologies, AI/ML techniques employed, key findings, and implications for SCRA. Data extraction forms were filled in to gather pertinent information for synthesis and further analysis. To comprehensively understand the research landscape, we conducted bibliometric and cocitation analyses of the 778 articles resulting from the filtering process. The bibliometric analysis helped us identify the field's most influential authors, journals, and key research themes. Cocitation analysis allowed us to identify clusters of related articles and determine their interconnections. Our study has several limitations that should be acknowledged. First, the search was limited to articles published between 2015 and 2025, and it is possible that relevant articles published before or after this timeframe were not included.

\section{Bibliometric Analysis}
\label{sec:biblio}
To comprehensively analyze the scholarly landscape of SCRA using AI, we employed a combination of tools, including MS Excel and VOSviewer. This bibliometric analysis encompasses several dimensions, offering a deeper understanding of the field by identifying trends, contributions, and intellectual structures.

\subsection{Publication Trend Over Time}

\begin{figure*}[!ht]
    \centering
    \includegraphics[width=0.8\textwidth]{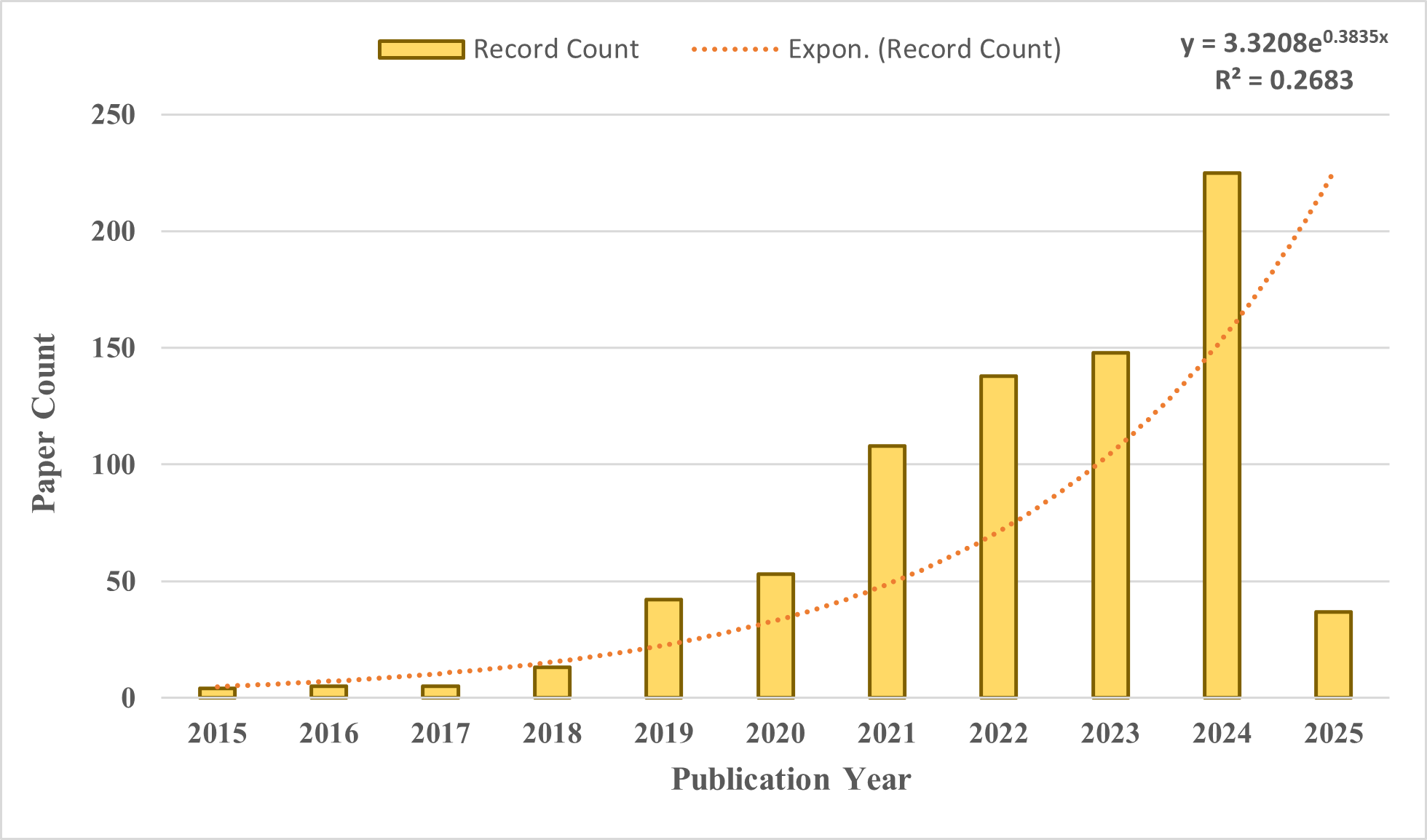}
    \caption{Annual publication trend (2015-2025) of the 778 published literature on AI in the SCRA topic.}
    \label{fig:publication_trend}
\end{figure*}

Figure~\ref{fig:publication_trend} illustrates the annual publication trends from 2015 to 2025. A significant rise in publications is observed post-2020, likely influenced by the COVID-19 pandemic, highlighting the critical need for resilient SC systems. The peak in research output during 2024 reflects the heightened importance of AI/ML-driven risk assessment in tackling complex SC challenges. However, the slight decline in 2025 is due to the first quarter, which is expected to peak following the exponential trend line.

\subsection{Geographical Distribution of Research}

\begin{figure*}[!ht]
    \centering
    \includegraphics[width=0.8\textwidth]{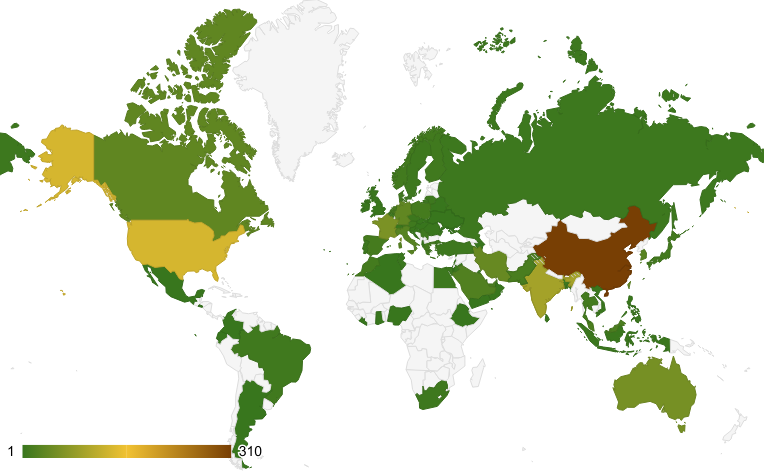}
    \caption{Geographical distribution of the authors affiliated with 778 published papers on AI in the SCRA topic.}
    \label{fig:geographical_distribution}
\end{figure*}

Figure~\ref{fig:geographical_distribution} visualizes the geographical distribution of research contributions. China leads with 39.49\% of 778 publications, followed by the United States (11.72\%) and England (11.59\%). This dominance underscores the strong focus of Chinese institutions on AI-driven SCRA. The global spread of contributions emphasizes the universal relevance of SCRA, with increasing participation from emerging economies.

\begin{table}[!ht]
\centering
\caption{Top 10 authors according to the published papers}
\label{tab:top_authors}
\footnotesize
%\resizebox{\linewidth}{!}{%
\begin{tblr}{
  column{2} = {c},
  hline{1-2,12} = {-}{},
}
\textbf{Authors} & \textbf{Published Paper Count} \\
Gupta S          & 14                              \\
Liu YK	         & 12                               \\
Brintrup A         &  11 \\
Kumar A          & 11                              \\
Mangla SK         & 8         \\
Modgil S         & 8                              \\
Choi TM         & 	7         \\
Liu Y         & 	7         \\
Li J         & 	6         \\
Liu C         & 	6         \\                          
\end{tblr}
%}
\end{table}
\normalsize

\subsection{Analysis of Authors}
As shown in Table \ref{tab:top_authors}, the top 10 authors have significantly influenced the domain of SCRA using AI/ML. Gupta S leads the field with 1.78\% of the 778 publications, followed by Liu YK (1.53\%) and Brintrup A (1.40\%), who have contributed substantially to advancing AI-driven risk assessment. Other key contributors include Kumar A (1.40\%), Mangla SK (1.02\%), and Modgil S (1.02\%), each demonstrating a strong research footprint in this domain. Additionally, Choi TM (0.89\%), Liu Y (0.89\%), Li J (0.76\%), and Liu C (0.76\%) round out the top 10 authors, further emphasizing the interdisciplinary and global nature of research in AI-based SCRA. These authors have played a crucial role in developing AI-driven risk assessment frameworks, predictive models, and decision-support systems, shaping the future of SC resilience and risk mitigation strategies.

\subsection{Publisher Contributions}

% Include figures or content about publisher contributions
\begin{figure*}[!ht]
    \centering
    \includegraphics[width=0.9\textwidth]{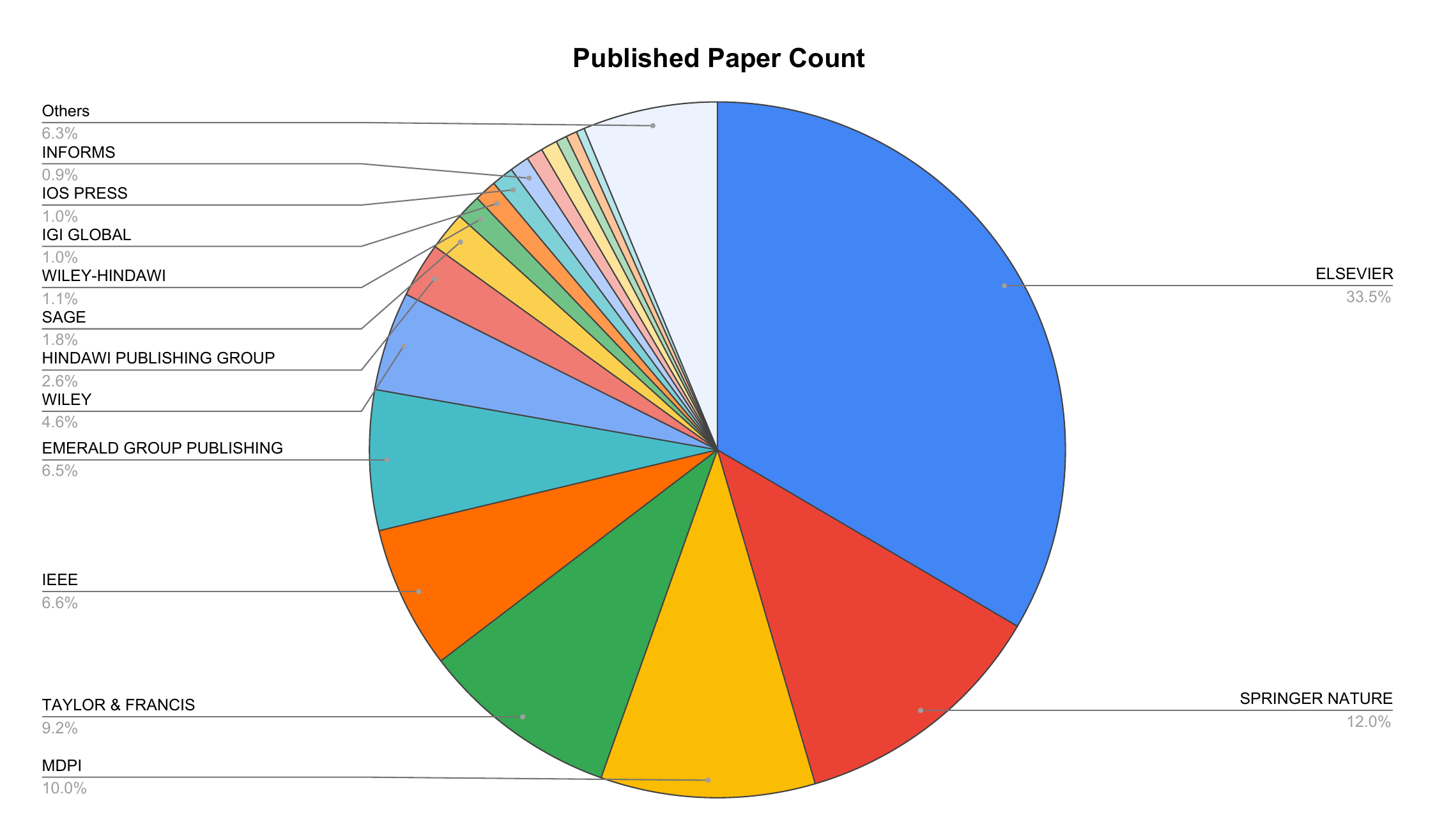}
    \caption{Publisher-wise distribution of 778 articles.}
    \label{fig:publisher_contribution}
\end{figure*}

Figure~\ref{fig:publisher_contribution} presents the contributions of various publishers. Elsevier is the most prominent, contributing over 262 articles, followed by Springer Nature, MDPI, Taylor \& Francis, and IEEE. These publishers serve as key platforms for disseminating high-quality research, reflecting their alignment with the evolving needs of AI and SCRA scholarship.

\begin{figure*}[!ht]
  \centering
  \begin{subfigure}{0.6\linewidth}
    \includegraphics[width=\linewidth]{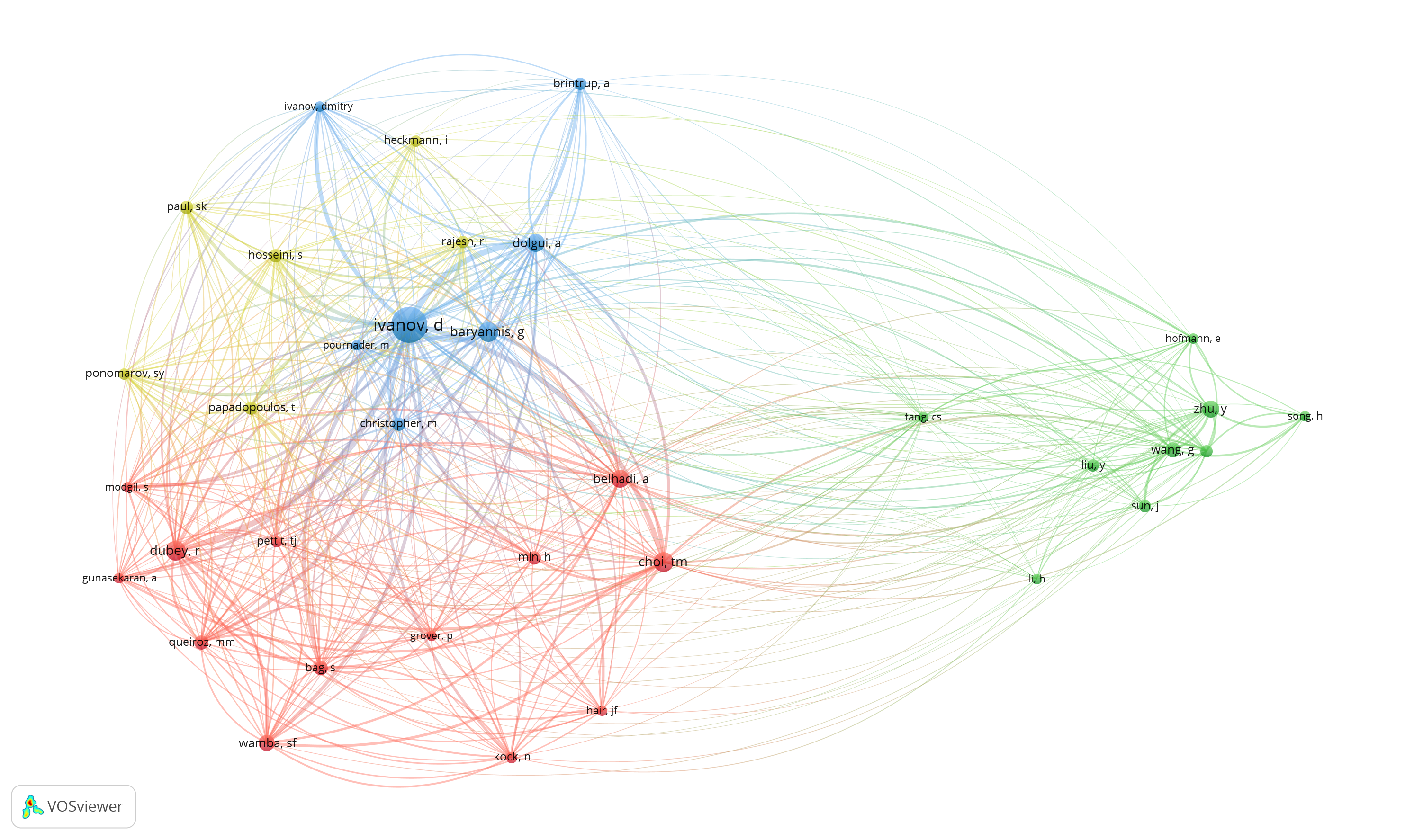}
    \caption{Author co-citations network graph.}
    \label{fig:cocitation_auth}
  \end{subfigure}
  \hfill
  \begin{subfigure}{0.6\linewidth}
    \includegraphics[width=\linewidth]{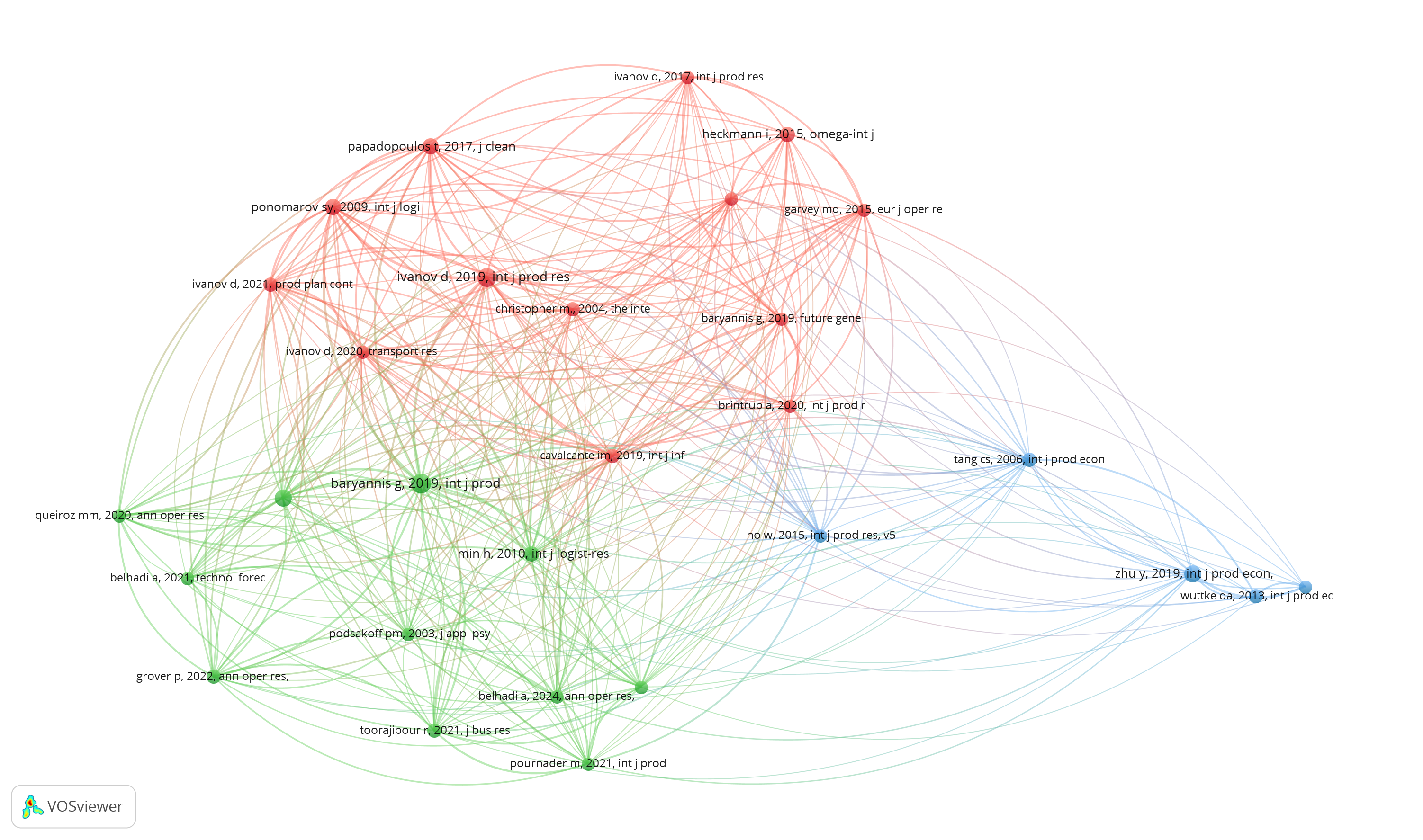}
    \caption{Co-citation network graph of cited references.}
    \label{fig:cocitation_ref}
  \end{subfigure}
    \hfill
  \begin{subfigure}{0.6\linewidth}
    \includegraphics[width=\linewidth]{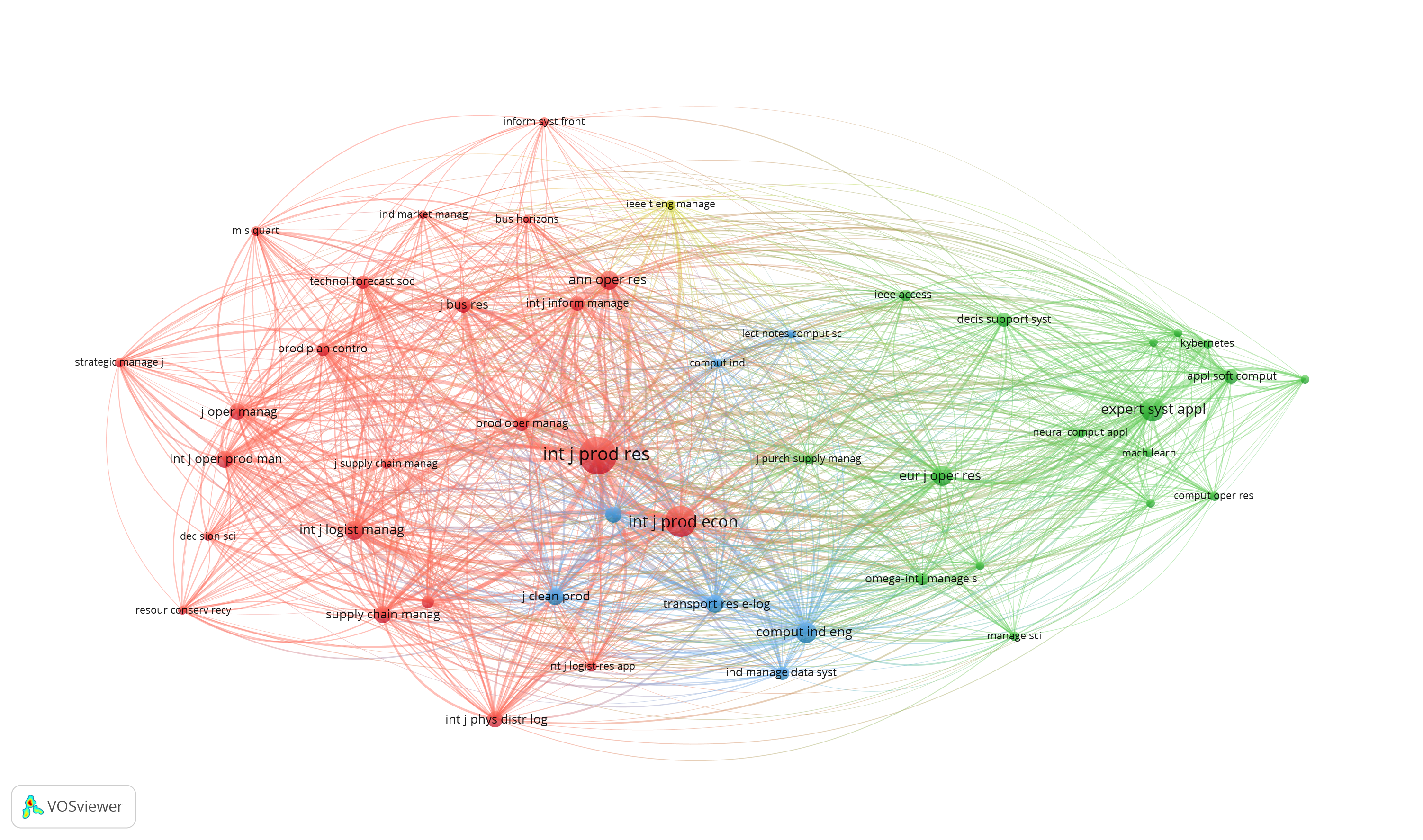}
    \caption{Co-citation network graph of publication sources.}
    \label{fig:cocitation_src}
  \end{subfigure}
  \caption{Co-citation network graphs visualization using (a) authors, (b) cited references, and (c) sources as the unit of analysis.}
  \label{fig:cocitation_main}
\end{figure*}

\subsection{Co-citation Analysis}
Co-citation analysis was conducted to explore the intellectual structure of AI-SCRA research, focusing on authors, references, and publication sources. Figure~\ref{fig:cocitation_auth} presents an author co-citation network comprising 55 authors grouped into three distinct clusters. The visualization highlights key contributors to AI-driven SCRA, with Ivanov and Baryannis playing central roles in the field. Other notable researchers, such as Dubey, Choi, and Zhu, also exhibit strong influence within their respective clusters. Figure~\ref{fig:cocitation_ref} presents a co-citation network with three main clusters, highlighting key contributions to AI-driven SCRM. Notable references, including Ivanov (2019, 2020), Baryannis (2019), and Tang (2006), emphasize hybrid AI models, risk prediction frameworks, and post-COVID resilience strategies. Lastly, Figure~\ref{fig:cocitation_src} underscores the centrality of academic journals within the co-citation network. The International Journal of Production Research (IJPR) emerges as the most influential venue, evidenced by its dominant node size and central positioning, which reflect high total link strength. Its proximity to the International Journal of Production Economics and Expert Systems with Applications highlights an intellectual core that bridges technical machine learning methodologies with operational management insights.

\begin{figure*}[!ht]
  \centering
  \begin{subfigure}{0.8\linewidth}
    \includegraphics[width=\linewidth]{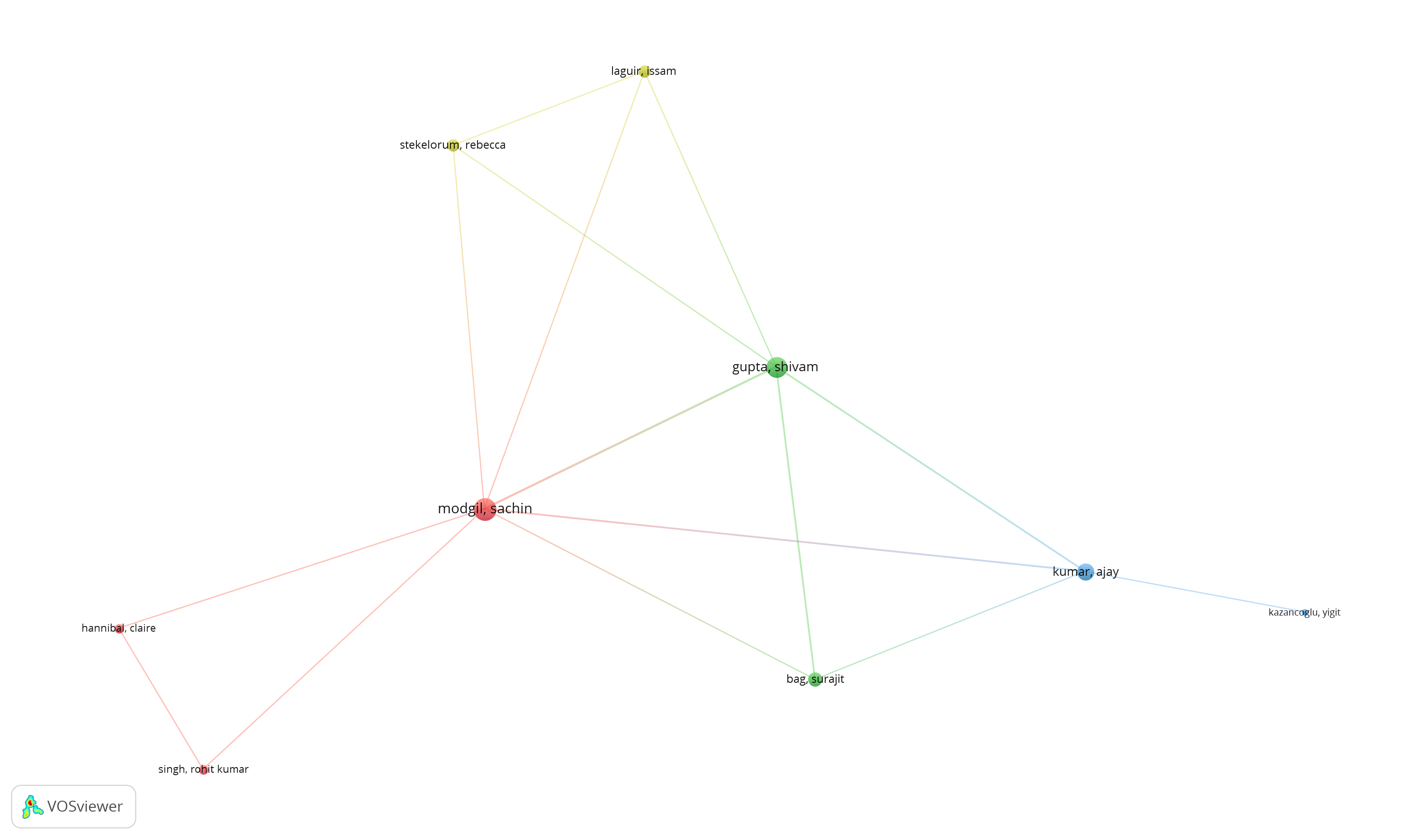}
    \caption{Co-authorship network analysis of 48 authors.}
    \label{fig:coauth_authors}
  \end{subfigure}
  \hfill
  \begin{subfigure}{0.8\linewidth}
    \includegraphics[width=\linewidth]{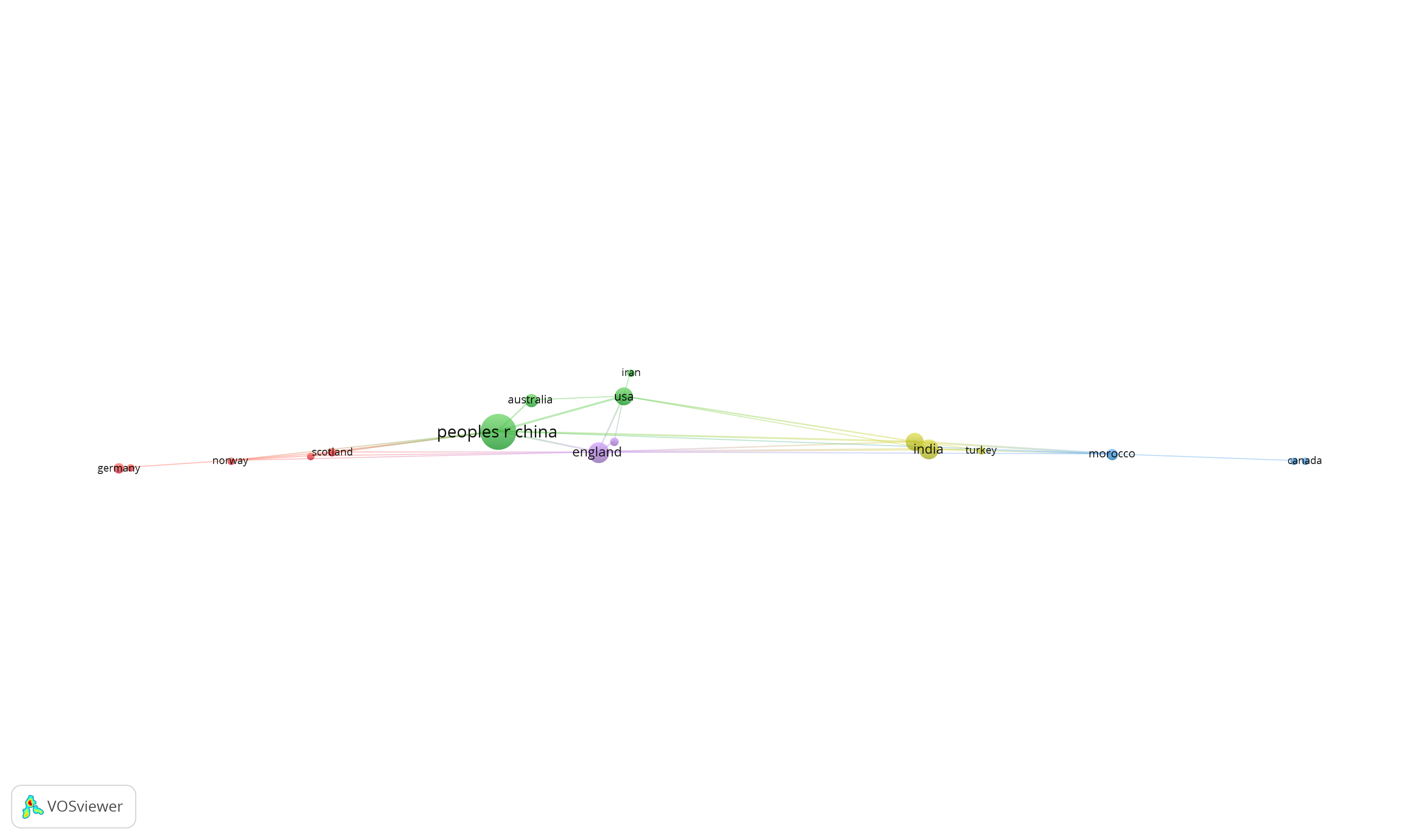}
    \caption{Co-authorship network analysis of 18 countries.}
    \label{fig:coauth_countries}
  \end{subfigure}
  \caption{Co-authorship network graphs visualizing collaborations at (a) author and (b) country levels.}
  \label{fig:co-authorship}
\end{figure*}

\subsection{Co-authorship Analysis}
Co-authorship analysis is a powerful method for understanding collaborations among authors and researchers through joint publications. This analysis highlights the collaborative nature of the research domain, uncovering influential partnerships and the significant contributions of specific countries. Figure~\ref{fig:coauth_authors} illustrates the co-authorship network among authors in this field. Using a threshold of at least one document and a minimum of 55 citations per author, 48 authors out of a total of 304 were identified as significant contributors. The visualization, generated using VOSviewer, reveals four distinct clusters, color-coded as red, green, blue, and yellow. These clusters represent groups of authors with strong collaborative ties. For example, the robust connections surrounding ``Modgil" in the red cluster indicate frequent co-authorship and shared research initiatives, reflecting a high degree of collaboration within the cluster. Figure~\ref{fig:coauth_countries} showcases the co-authorship network at the country level. Applying thresholds of at least 2 documents and 9 citations per country, 18 of 36 countries were included in the analysis. The visualization highlights China, the USA, and England as leading contributors, dominating document count, citation impact, and total link strength. This dominance underscores their pivotal roles in advancing AI/ML-driven SCRA research. The results also emphasize strong international collaborations between these countries and other key players, fostering a globally interconnected research ecosystem.

\begin{figure*}[!ht]
  \centering
  \begin{subfigure}{0.8\linewidth}
    \includegraphics[width=\linewidth]{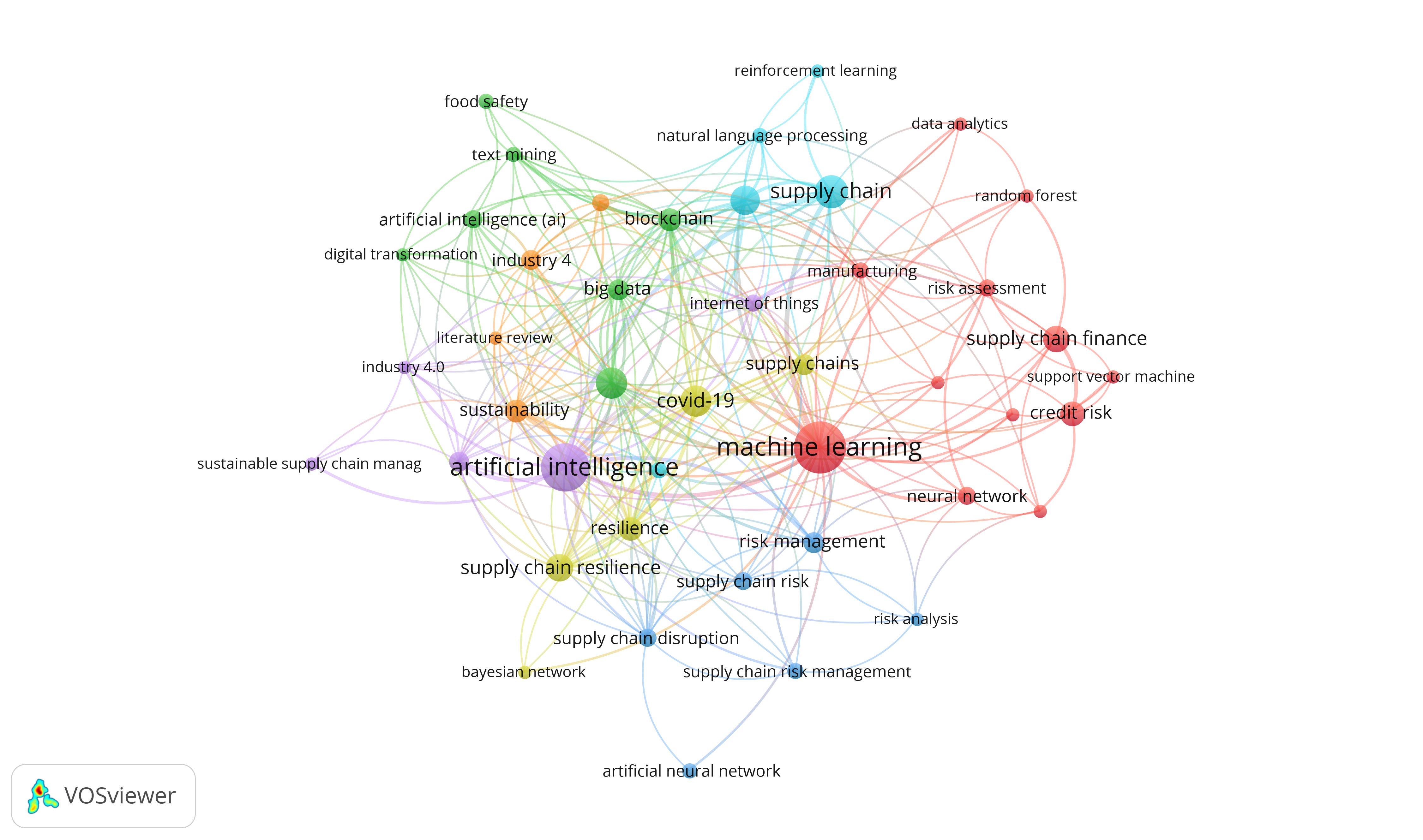}
    \caption{Co-occurrence network analysis of top 26 authors’ keywords.}
    \label{fig:authors_key}
  \end{subfigure}
  \hfill
  \begin{subfigure}{0.8\linewidth}
    \includegraphics[width=\linewidth]{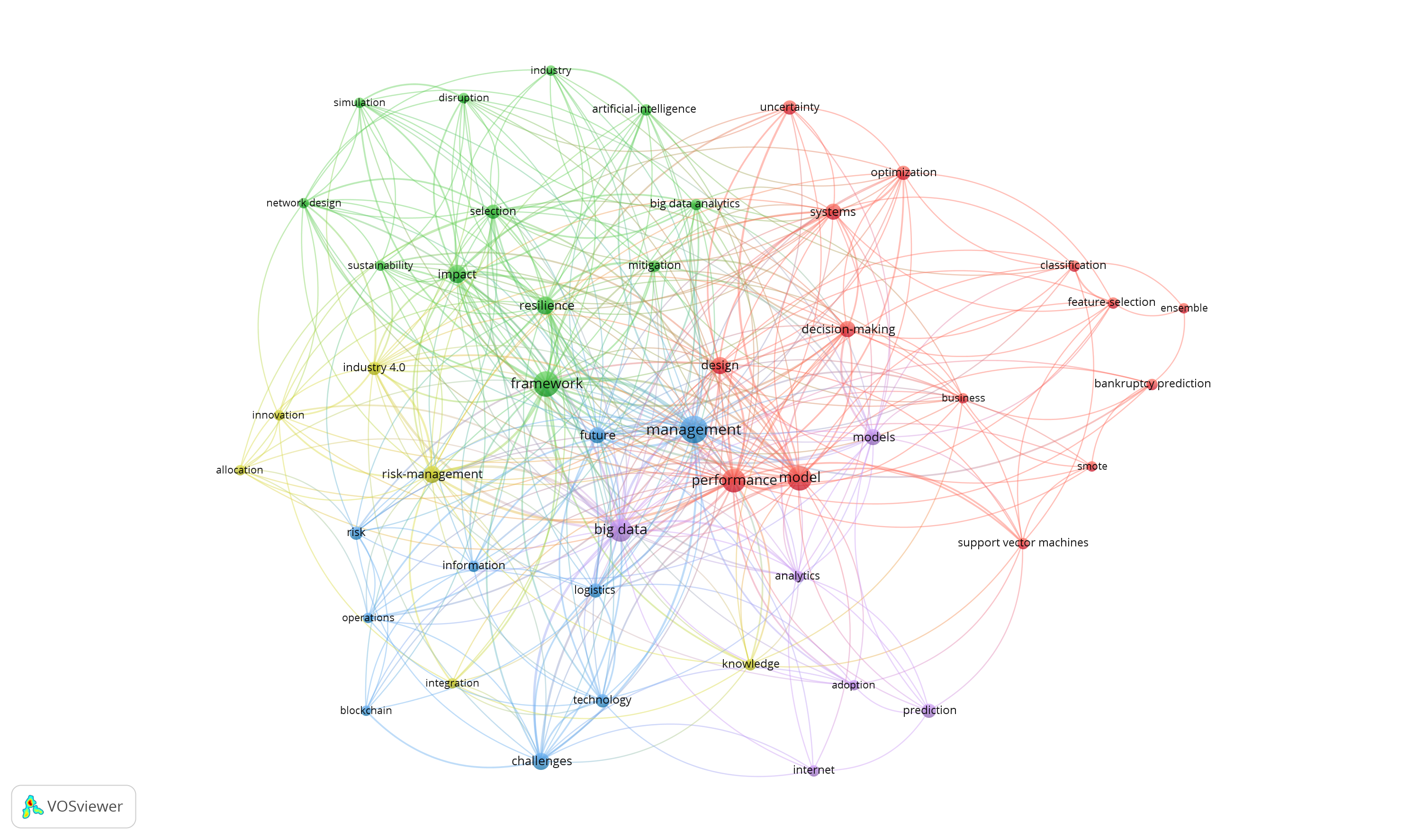}
    \caption{Co-occurrence network analysis of top 47 keywords used in the literature related to SCRA-AI.}
    \label{fig:key_occurrence}
  \end{subfigure}
  \caption{The Co-occurrence Network visualizes the intricate relationships among identified keywords. Each node represents a keyword, and the edges denote the frequency of co-occurrence. Clusters of tightly connected nodes signify thematic groupings, providing a comprehensive snapshot of the intellectual structure within the analyzed literature.}
  \label{fig:co-occurrence}
\end{figure*}

\subsection{Co-occurrence Analysis}
The co-occurrence analysis offers valuable insights into the thematic structure of SCRA and AI/ML research by identifying how keywords are interconnected within the scholarly literature. Using VOSviewer, we conducted this analysis by focusing on the units of analysis, such as ``Authors' Keywords" and ``Keywords Plus." This approach reveals latent connections and thematic clusters, enabling a deeper understanding of this domain's critical topics and prevalent themes. The network map shown in Figure~\ref{fig:authors_key} illustrates the relationships among the 26 most frequently occurring author keywords. These keywords were selected from an initial pool of 1,483, applying a minimum threshold of three occurrences. Larger nodes, such as ``artificial intelligence" and ``machine learning," indicate higher frequencies. The proximity of the nodes reflects their degree of association, with closely grouped nodes forming thematic clusters. Four clusters emerged, highlighting key themes. Figure~\ref{fig:key_occurrence} illustrates the co-occurrence of keywords in the reviewed literature. The analysis identified 47 frequently occurring terms from 1,032, using a minimum threshold of three occurrences. These keywords were categorized into five distinct clusters, reflecting prevalent themes and connections within the research field. Terms such as ``management" and ``framework" exhibited the strongest linkages, indicating their crucial roles in connecting different thematic areas. Additionally, other significant clusters highlighted emerging trends, including ``supply chain resilience", ``big data", ``blockchain", and ``sustainability". This co-occurrence analysis offers a comprehensive overview of the intellectual structure of SCRA research, showcasing established and emerging themes. The findings emphasize the interconnectivity of concepts and provide guidance for future research directions by identifying critical areas of scholarly focus.

\begin{table*}[!ht]
\centering
\caption{Top 10 Affiliations According to Paper Count}
\label{tab:top_affiliations}
\footnotesize
% \resizebox{\textwidth}{!}{%
\begin{tabular}{lcc}
\hline
\textbf{Affiliations}                                     & \textbf{Published paper count} & \textbf{\% of 778 articles} \\ \hline
HONG KONG POLYTECHNIC UNIVERSITY                 & 22                    & 2.799                       \\
CHINESE ACADEMY OF SCIENCES                      & 21                    & 2.672                       \\
INDIAN INSTITUTE OF TECHNOLOGY (IIT) SYSTEM & 20                    & 2.545                       \\
UNIVERSITY OF TEHRAN                             & 19                    & 2.417                       \\
INDIAN INSTITUTE OF MANAGEMENT (IIM) SYSTEM        & 15                    & 1.908                       \\
NEOMA BUSINESS SCH                               & 15                    & 1.908                       \\
UNIVERSITY OF CAMBRIDGE                          & 15                    & 1.908                       \\
HEBEI UNIVERSITY                                 & 14                    & 1.781                       \\
EMLYON BUSINESS SCHOOL                           & 11                    & 1.399                       \\
INDIAN INSTITUTE OF TECHNOLOGY (IIT) DELHI         & 11                    & 1.399                       \\
Others                                           & 615                   & 79.26                       \\ \hline
\end{tabular}%
% }
\end{table*}
\normalsize

\subsection{Top 10 Affiliations}
Table \ref{tab:top_affiliations} highlights the top 10 affiliations contributing to SCRA research, ranked by their published paper count. Hong Kong Polytechnic University leads the list with 22 publications (2.80\%), followed closely by the Chinese Academy of Sciences (2.67\%) and the Indian Institute of Technology (IIT) System (2.55\%). Other key contributors include the University of Tehran (2.42\%), the Indian Institute of Management (IIM) System (1.91\%), and NEOMA Business School (1.91\%), all of which have played a crucial role in advancing AI-driven SCRA methodologies. Prominent European institutions such as the University of Cambridge (1.91\%), Hebei University (1.78\%), and Emlyon Business School (1.40\%) further highlight the global nature of SCRA research, emphasizing interdisciplinary contributions across Asia, Europe, and the Middle East. Notably, IIT Delhi (1.40\%) is a significant research hub in this field. Together, these top 10 affiliations account for a substantial share of high-impact studies in SCRA, driving advancements in AI applications for risk mitigation. Meanwhile, the ``Others" category, comprising 615 institutions, represents 79.26\% of the contributions, reflecting the diverse and widespread nature of research in this domain.

\subsection{Bibliographic Coupling}
Bibliographic coupling assesses the similarity of documents, authors, or organizations based on shared references, revealing thematic alignments and intellectual structures. Key insights are as follows:

\subsubsection{Authors}
\begin{figure*}[!ht]
    \centering
    \includegraphics[width=0.9\linewidth]{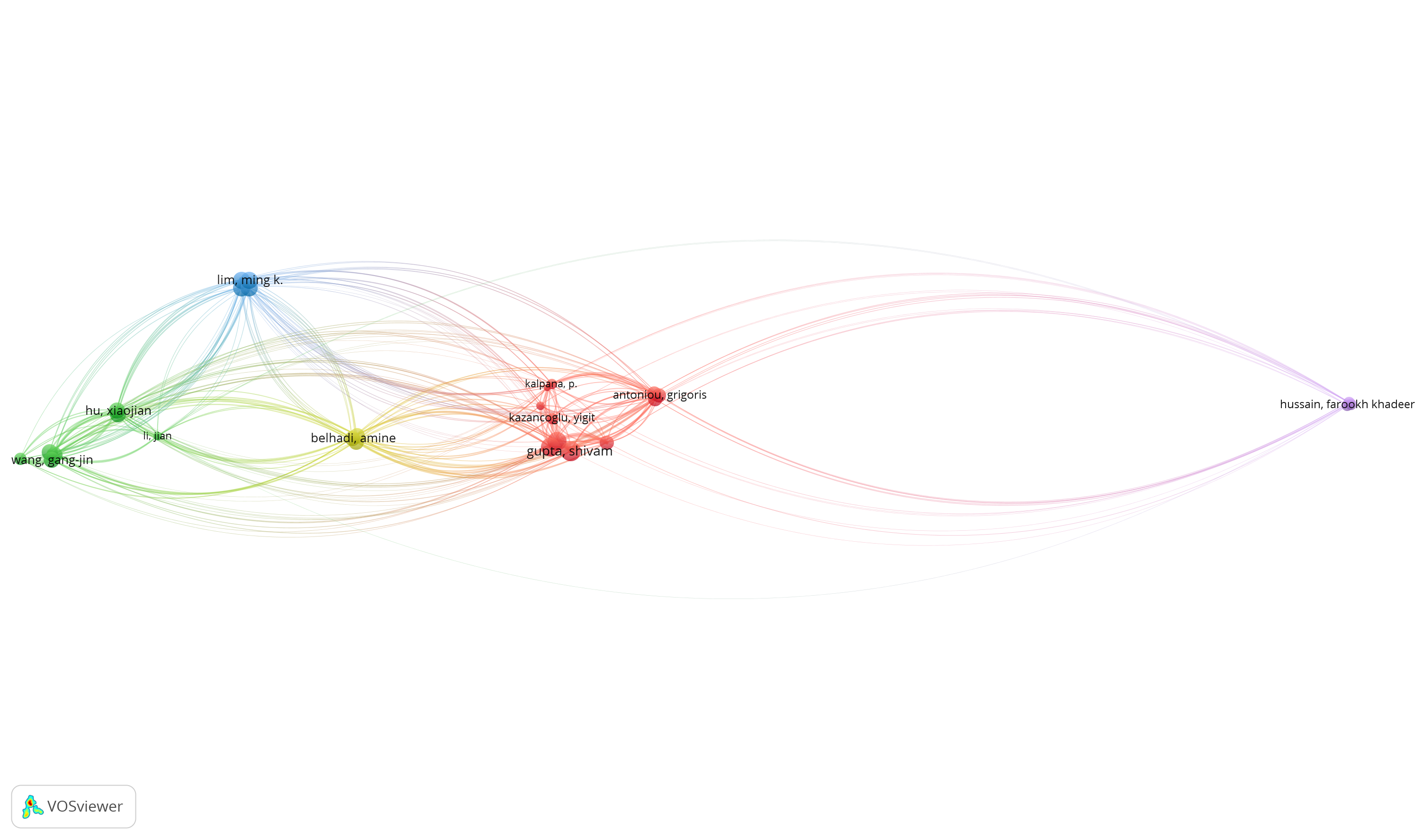}
    \caption{Bibliographic coupling analysis of 28 authors.}
    \label{fig:bib_auth}
\end{figure*}

Figure~\ref{fig:bib_auth} illustrates the bibliographic coupling analysis of 28 authors, grouping them into five thematic clusters based on shared references, highlighting intellectual connections in SCR research. Prominent contributors like Lim M.K., Gupta, Shivam, Wang G.-J., and Hussain F.K. demonstrate centrality, reflecting their significant influence in areas such as risk prediction, disruption management, and quantitative modeling. The clusters reveal interdisciplinary themes, including resilience strategies, industry-specific applications, and the integration of AI-driven methods. This analysis showcases established research directions and emerging trends, providing a roadmap for future studies and collaborations in SCR.

\subsubsection{Countries}
\begin{figure*}[!ht]
    \centering
    \includegraphics[width=0.7\textwidth]{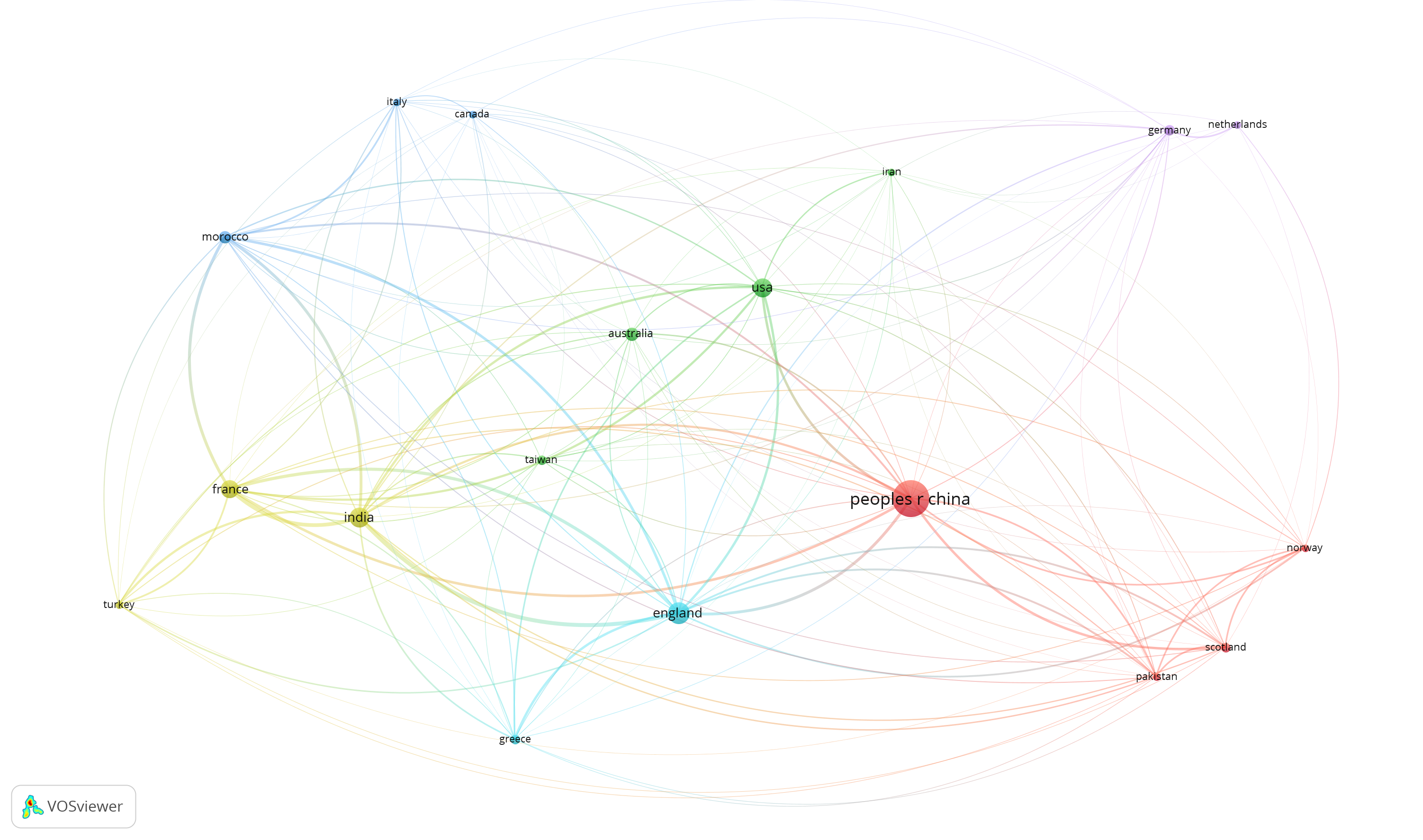}
    \caption{Bibliographic coupling analysis of 18 countries.}
    \label{fig:bib_countries}
\end{figure*}

Figure~\ref{fig:bib_countries} illustrates the bibliographic coupling among 18 countries, categorizing them into clusters highlighting global research collaborations in AI-driven SCRA. Countries such as China, the United States, and England emerge as central hubs, showcasing their pivotal roles in advancing research and promoting international partnerships. The clusters signify thematic alignments across regions, with nations collaborating on critical topics like predictive analytics, sustainability, and resilience strategies.

\subsubsection{Documents}
\begin{figure*}[!ht]
    \centering
    \includegraphics[width=0.7\textwidth]{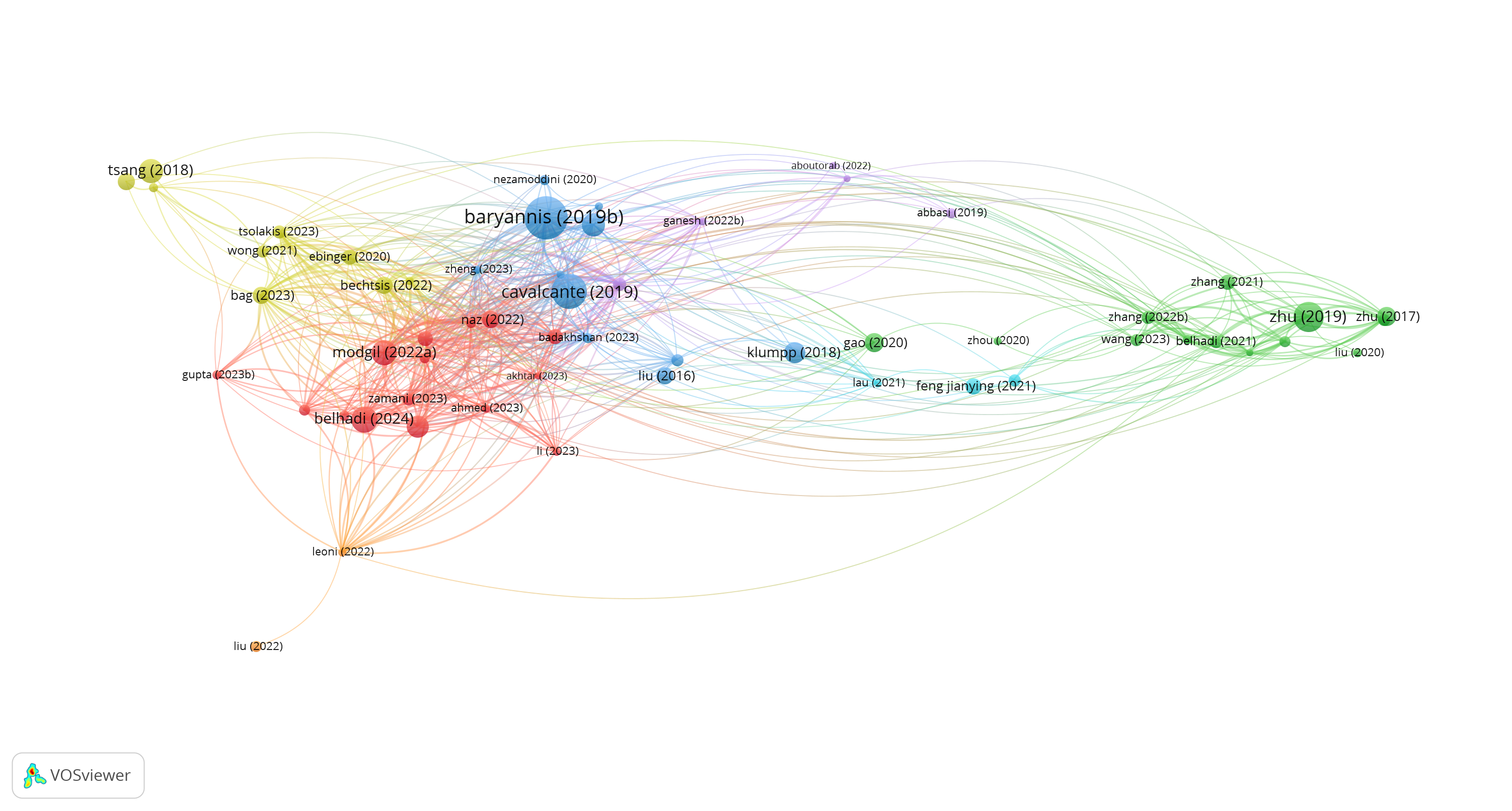}
    \caption{Bibliographic coupling analysis of 62 documents.}
    \label{fig:bib_doc}
\end{figure*}

Figure~\ref{fig:bib_doc} illustrates the bibliographic coupling analysis of 62 documents, visually depicting the thematic clustering and citation relationships among the studies. The eight clusters highlight distinct research themes within the field, showcasing the interconnectedness of key works. Highly influential documents, such as those by ``Baryannis G.", ``Cavalcante I. M.", and ``Zhu Q.", emerge as central nodes, reflecting their significant contributions to the discourse on SCRA using AI. This visualization underscores the collaborative and evolving nature of the research domain, providing insights into how various studies align and contribute to specific thematic advancements.

\subsubsection{Organizations}
\begin{figure*}[!ht]
    \centering
    \includegraphics[width=0.7\textwidth]{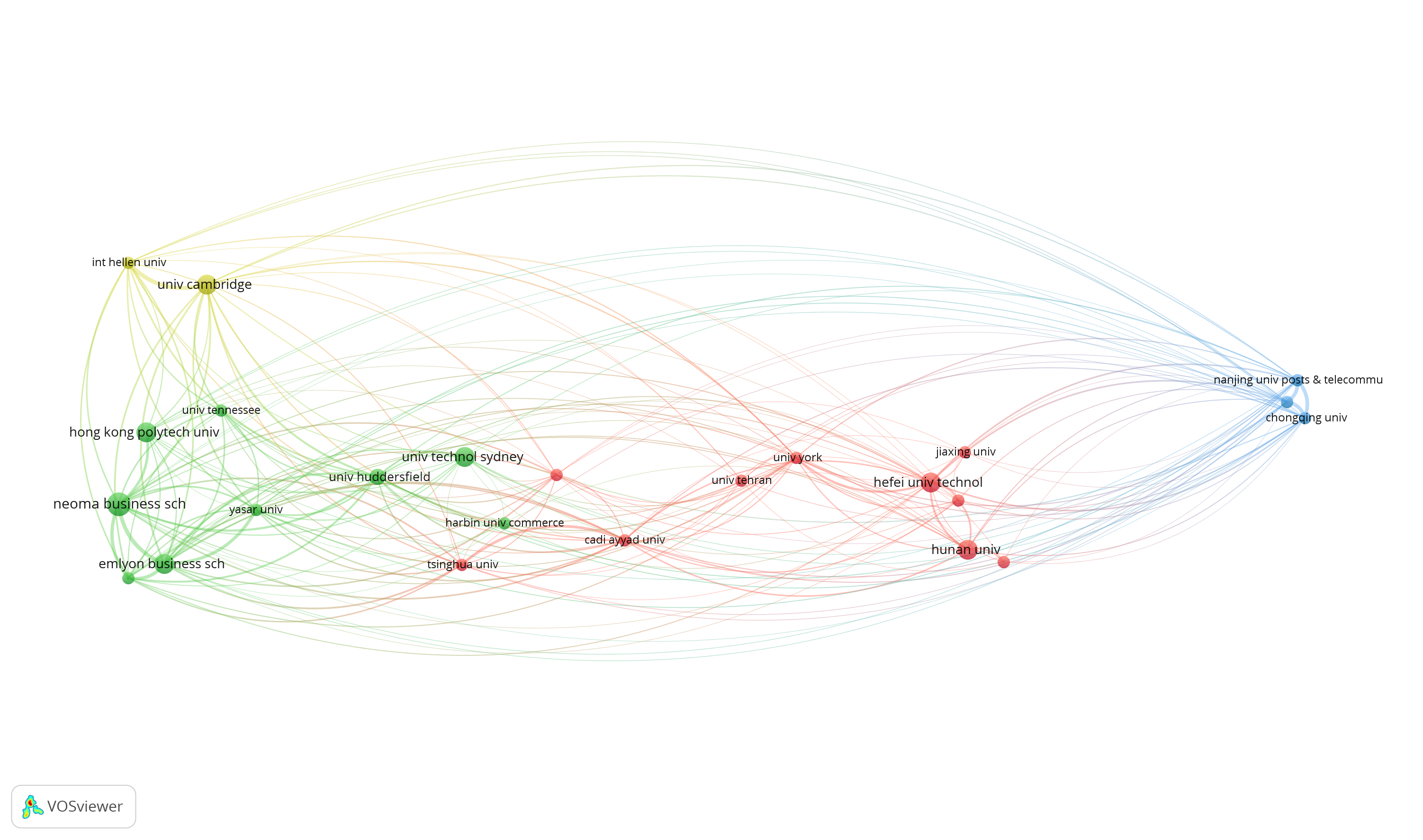}
    \caption{Bibliographic coupling analysis of 24 organizations.}
    \label{fig:bib_org}
\end{figure*}

Figure~\ref{fig:bib_org} presents the bibliographic coupling analysis of 24 organizations, highlighting their collaborative and thematic alignments in AI-based SCRA research. The visualization categorizes institutions into three clusters, highlighting key contributors such as the Hong Kong Polytechnic University, the University of Cambridge, and Nanjing University of Posts and Telecommunications. This analysis underscores the global distribution of research efforts and the leadership roles of prominent institutions in advancing AI-driven methodologies for SCRA.

\subsubsection{Journals}
\begin{figure*}[!ht]
    \centering
    \includegraphics[width=0.7\textwidth]{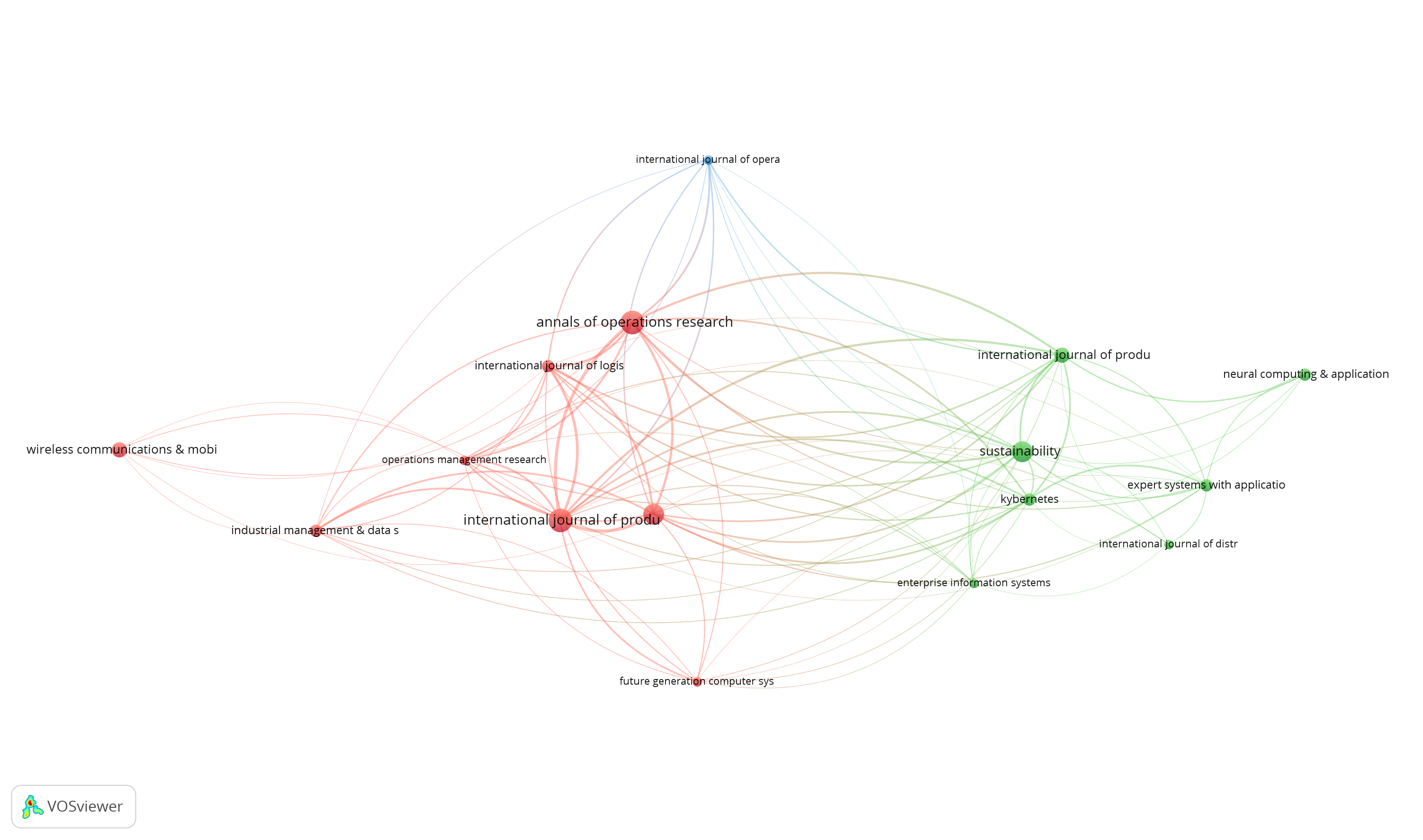}
    \caption{Bibliographic coupling analysis of 17 publication sources.}
    \label{fig:bib_src}
\end{figure*}

Figure~\ref{fig:bib_src} illustrates the bibliographic coupling analysis of the core academic journals driving the advancements in SCRA research by showing the interconnections and thematic clustering among 17 influential journals. Journals such as the ``International Journal of Production Research" and ``Annals of Operations Research" play a crucial role, providing foundational studies and methodologies that inform AI-driven approaches to SCRA. These clusters reveal thematic trends and emerging focus areas and facilitate knowledge dissemination by connecting researchers across disciplines.

\begin{table*}[!ht]
\caption{Publication Trends by Journal (2015-2025)}
\label{tab:publication_trends}
\resizebox{\textwidth}{!}{%
\begin{tabular}{@{}lcccccccccccc@{}}
\toprule
\multirow{2}{*}{\textbf{Source   Title}}                                 & \multicolumn{11}{c}{\textbf{Publication Year}}                                                                                                                                & \multirow{2}{*}{\textbf{Grand Total}} \\ \cmidrule(lr){2-12}
                                                                         & \textbf{2015} & \textbf{2016} & \textbf{2017} & \textbf{2018} & \textbf{2019} & \textbf{2020} & \textbf{2021} & \textbf{2022} & \textbf{2023} & \textbf{2024} & \textbf{2025} &                                       \\ \midrule
INTERNATIONAL JOURNAL OF PRODUCTION   RESEARCH                           & 1             &               &               &               &               & 5             &               & 3             & 4             & 18            & 7             & 38                                    \\
SUSTAINABILITY                                                           &               & 1             &               &               & 2             & 4             & 5             & 9             & 9             & 3             & 1             & 34                                    \\
ANNALS OF OPERATIONS RESEARCH                                            &               &               &               &               &               &               & 2             &               & 10            & 18            & 1             & 31                                    \\
ENGINEERING APPLICATIONS OF ARTIFICIAL   INTELLIGENCE                    & 1             &               &               &               & 1             & 2             & 2             &               & 5             & 8             & 2             & 21                                    \\
INTERNATIONAL JOURNAL OF PRODUCTION   ECONOMICS                          &               & 1             &               &               & 1             & 1             & 1             & 1             & 6             & 8             & 2             & 21                                    \\
COMPUTERS \& INDUSTRIAL ENGINEERING                                      &               &               &               & 2             & 1             &               & 3             & 2             & 3             & 5             & 4             & 20                                    \\
IEEE ACCESS                                                              &               &               &               &               & 2             &               & 4             & 1             & 2             & 4             & 3             & 16                                    \\
IEEE TRANSACTIONS ON ENGINEERING   MANAGEMENT                            &               &               &               &               &               &               &               &               & 4             & 12            &               & 16                                    \\
EXPERT SYSTEMS WITH APPLICATIONS                                         &               & 1             &               &               & 1             &               & 1             & 2             & 3             & 4             & 3             & 15                                    \\
TRANSPORTATION RESEARCH PART E-LOGISTICS   AND TRANSPORTATION REVIEW     &               &               &               &               & 1             &               & 1             & 1             & 3             & 1             & 4             & 11                                    \\
JOURNAL OF CLEANER PRODUCTION                                            &               &               &               &               &               & 2             &               & 5             & 1             & 2             &               & 10                                    \\
APPLIED SCIENCES-BASEL                                                   &               &               &               &               &               &               & 1             & 2             & 1             & 5             &               & 9                                     \\
INDUSTRIAL MANAGEMENT \& DATA SYSTEMS                                    &               &               &               & 1             &               &               & 3             & 1             & 2             &               &               & 7                                     \\
JOURNAL OF INTELLIGENT \& FUZZY   SYSTEMS                                &               &               &               &               & 1             & 2             & 3             &               & 1             &               &               & 7                                     \\
PRODUCTION AND OPERATIONS MANAGEMENT                                     &               &               &               &               &               &               &               & 4             &               & 1             & 2             & 7                                     \\
TECHNOLOGICAL FORECASTING AND SOCIAL   CHANGE                            &               &               &               & 1             &               &               &               & 2             & 2             & 2             &               & 7                                     \\
APPLIED SOFT COMPUTING                                                   &               &               & 1             &               &               & 1             &               &               & 1             & 1             & 2             & 6                                     \\
DECISION SUPPORT SYSTEMS                                                 &               &               &               &               &               & 1             &               & 3             &               & 2             &               & 6                                     \\
EUROPEAN JOURNAL OF OPERATIONAL RESEARCH                                 &               &               &               &               & 1             & 1             & 2             & 1             &               &               & 1             & 6                                     \\
INDUSTRIAL MARKETING MANAGEMENT                                          &               &               &               &               &               & 1             & 1             & 1             &               & 3             &               & 6                                     \\
INTERNATIONAL JOURNAL OF   LOGISTICS-RESEARCH AND APPLICATIONS           &               &               &               &               &               &               &               &               & 3             & 3             &               & 6                                     \\
INTERNATIONAL JOURNAL OF PHYSICAL   DISTRIBUTION \& LOGISTICS MANAGEMENT &               &               &               &               &               &               & 1             & 2             & 1             & 2             &               & 6                                     \\
OPERATIONS MANAGEMENT RESEARCH                                           &               &               &               &               &               &               &               & 3             & 1             & 2             &               & 6                                     \\
PLOS ONE                                                                 &               &               &               &               & 1             &               & 1             & 1             & 2             & 1             &               & 6                                     \\
WIRELESS COMMUNICATIONS \& MOBILE   COMPUTING                            &               &               &               &               &               &               & 3             & 3             &               &               &               & 6                                     \\
Others                                                                   & 2             & 2             & 4             & 9             & 27            & 28            & 46            & 77            & 68            & 151           & 40            & 454                                   \\
\textbf{Grand Total}                                                     & \textbf{4}    & \textbf{5}    & \textbf{5}    & \textbf{13}   & \textbf{39}   & \textbf{48}   & \textbf{80}   & \textbf{124}  & \textbf{132}  & \textbf{256}  & \textbf{72}   & \textbf{778}                          \\ \bottomrule
\end{tabular}%
}
\end{table*}

\subsection{Journal Publication Trends}
Table~\ref{tab:publication_trends} presents a comprehensive breakdown of published papers in various source titles from 2015 to 2025. These source titles encompass a range of journals and publications contributing to the field of SCRA. The table clearly shows the publication trends, indicating how many papers were published in each source title each year and their cumulative total. It provides valuable insights into the distribution of research output and the source titles active in this area.

\section{Framework for SCRA using AI}
\label{sec:Framework for SCRA using AI}

\begin{figure*}[!ht]
    \centering
    \includegraphics[width=1\textwidth]{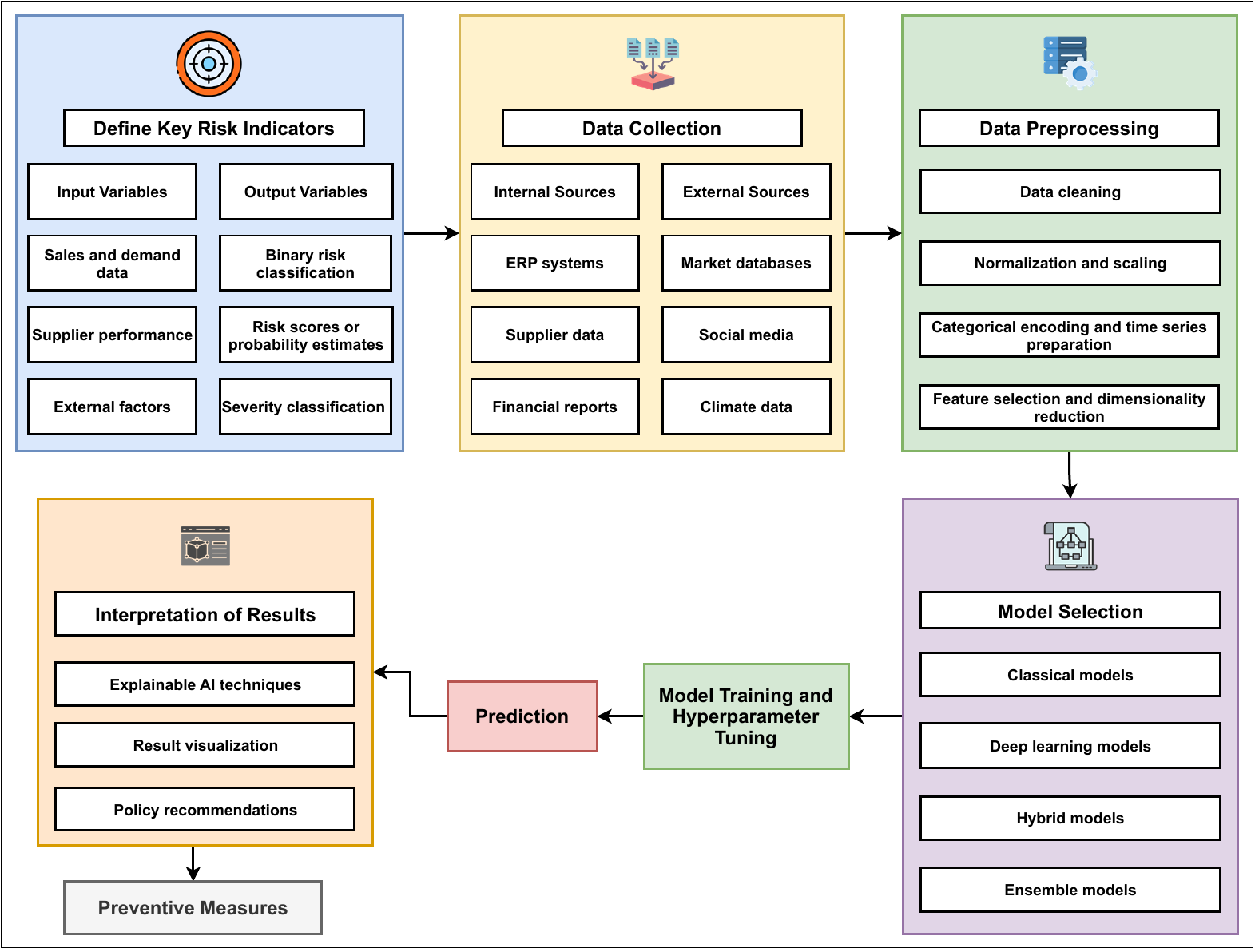}
    \caption{Framework for SCRA using AI.}
    \label{fig:framework}
\end{figure*}

Our proposed AI-driven SCRA framework provides a comprehensive pipeline for SC professionals to effectively integrate AI techniques. Figure~\ref{fig:framework} visually shows the sequential steps involved, from defining key risk indicators to implementing preventive measures. Following these steps, practitioners can systematically define risk indicators, source and pre-process data, select and train models, and interpret results to facilitate informed decision-making. The framework is enriched with insights from existing studies on AI and ML applications in SC risk, presenting best practices, key considerations, and real-world challenges and opportunities identified in the literature.

\subsection{Defining Key Risk Indicators}  
Our first step is identifying the key input and output variables that drive SCRA. Input variables include transactional data, supplier metrics, and external factors like political or environmental conditions, while output variables classify risks into severity levels such as ``low," ``medium," and ``high" to align with organizational decision-making priorities. Defining these variables is crucial, as they represent the foundation upon which AI models assess risk. \cite{wu_text-based_2024} emphasizes capturing operational and risk-related data for comprehensive risk assessments, with \cite{xu_enterprise_2024} using historical sales and market trends for disruption prediction. Including financial and non-financial indicators, as highlighted by \cite{wang_multiview_2021}, improves the assessment of risk exposure. Practitioners should ensure these variables align with their industry’s risk characteristics to yield actionable, context-relevant insights.

\subsection{Data Collection}  
Effective risk assessment relies on internal and external data sources to capture a comprehensive picture of potential risks. Internal data from ERP systems, procurement records, and inventory management systems provides insights into operational processes. External data, such as market indicators, social media sentiment, and climate reports, are critical for \cite{deiva_ganesh_supply_2022}, which used real-time data from Twitter as the input source. ERP systems, supplier systems, and logistics systems, alongside external sources like social media and weather forecasts. Such data covers both structured (e.g., numerical) and unstructured (e.g., text or multimedia) formats to improve risk detection. \cite{deiva_ganesh_supply_2022} used real-time data from social media, that is, Twitter, as the input source. The author used the Twitter streaming API as a primary source for data extraction. \cite{kosasih_towards_2022} used external platforms such as Marklines for automotive SC data and Achilles Information Ltd for energy sector insights. \cite{pan_dynamics_2023} collected data through surveys from industry practitioners to capture qualitative and contextual insights about SCR.Including such data enhances the model’s ability to preemptively detect risks, such as demand fluctuations or reputational issues, underscoring the value of a multimodal data approach.

\subsection{Data Preprocessing}  
Preprocessing is a vital step. Raw data must undergo preprocessing to improve model accuracy and efficiency. This step involves cleaning data to remove noise, normalizing values, and encoding categorical variables. Dimensionality reduction techniques like principal component analysis (PCA) applied by \cite{luo_construction_2022} and imputation methods like residual mean replacement applied by \cite{zhang_firefly_2021} address challenges such as high-dimensional datasets and missing values. Time-series data, common in SC applications, requires additional adjustments, such as seasonality corrections. Techniques such as Isolation Forest and Local Outlier Factor (LOF) detect anomalies, thereby enhancing prediction reliability. SPSS statistical preprocessing methods have also been employed effectively by \cite{dang_evaluating_2022}.

\subsection{Model Selection}  
Practitioners have a range of model choices, each suited to different data characteristics and objectives in SCRA. Classical models like Support Vector Machine (SVM), Decision Tree (DT), and Logistic Regression (LR) are ideal for well-structured, low-dimensional datasets, where the data relationships are more straightforward and computational efficiency is prioritized. DL models, such as Convolutional Neural Networks (CNNs) and Long Short-Term Memory (LSTM) networks, are valuable for handling complex, high-dimensional data, including time-series demand data, where capturing temporal dependencies is essential. Hybrid models, which combine conventional ML and DL approaches, balance interpretability and accuracy, catering to scenarios where insights into model decisions are as important as predictive performance. Ensemble methods like Random Forest (RF) and XGBoost aggregate outputs from multiple models to improve predictive accuracy and robustness. Additional research emphasizes that model selection should consider data availability, computational resources, and the necessity for model interpretability. Highly interpretable models are particularly important in regulated industries, such as finance, where transparency is critical. Hybrid and ensemble models, increasingly popular in recent literature, often effectively compromise predictive power and interpretability, making them suitable for complex SC risk scenarios.

\subsection{Model Training and Hyperparameter Tuning}  
Model training typically involves cross-validation to improve model robustness. Tuning parameters such as learning rate, depth, and the number of estimators (in ensemble methods) is crucial for optimizing performance. Tailored hyperparameter tuning can significantly improve model performance in SC applications. Many researchers highlight the importance of automated tuning techniques, such as grid search, random search, and Bayesian optimization, to streamline model refinement and ensure that the final model is well-suited for specific risk scenarios. \cite{wu_analysis_2022} emphasized iterative optimization, while \cite{zhang_credit_2022} addressed data imbalance with SMOTENC. Practitioners may also consider transfer learning to leverage knowledge from similar datasets, especially when labeled data is limited.

\subsection{Prediction}  
Once trained, the models generate predictions in the form of classifications (e.g., ``high risk" vs. ``low risk"), severity scores, or probabilities, which guide risk mitigation strategies. \cite{baryannis_supply_2019} highlighted the role of predictive analytics in shielding SCs from disruptions. These predictions should ideally be integrated into real-time dashboards to enable continuous monitoring and provide quick insights into risk fluctuations. This step facilitates proactive risk management, enabling organizations to adjust strategies promptly based on predicted risk levels and preventing disruptions before they escalate.

\subsection{Interpretation of Results}  
Interpretability makes AI-driven predictions actionable, especially for stakeholders without a technical background. XAI tools, such as SHAP values or LIME, help practitioners understand which variables most influence predictions. According to \cite{olan_enabling_nodate}, techniques such as SHAP and LIME improve decision-making by elucidating the rationale behind ML predictions, thereby enhancing interpretability and fostering trust in SCRA. Interpretation tools also validate results with domain experts, ensuring the model's output aligns with real-world SC conditions and providing a basis for actionable policy recommendations.

\subsection{Preventive Measure}  
Based on the risk predictions and insights, practitioners can establish preventive strategies to mitigate potential SC disruptions. Common actions include adjusting inventory buffers, diversifying suppliers, or modifying production schedules. Preventive measures should be continuously refined based on AI predictions and risk evaluations. Incorporating feedback loops that allow practitioners to assess the impact of preventive measures on risk reduction is a good practice. Such loops can help organizations refine their approach over time, ensuring that the AI-driven risk assessment framework remains adaptable to changing conditions.

% \onecolumn 
\footnotesize
\begin{longtable}[!ht]{p{3cm}p{5.7cm}p{6.5cm}}
\caption{Methods used and the key findings from the review articles.}
\label{tab:ml_methods}\\

\hline
\textbf{Reference(s)} & \textbf{Methods/Models} & \textbf{Key Findings} \\
\hline
\endfirsthead

\caption*{\textbf{Table \ref{tab:ml_methods} (Continued):} Methods used and the key findings from the review articles.}\\
\hline
\textbf{Reference(s)} & \textbf{Methods/Models} & \textbf{Key Findings} \\
\hline
\endhead 

\cite{katsaliaki_supply_2022} & Tensorization-based neural symbolic model & Outperformed SNLP and GNN-LP in predicting relationships (AUC: 0.82, AP: 0.72). \\ 
\hline
\cite{wang_forecasting_2022} & Cost-sensitive learning to a RF (CSL-RF) model & Improved prediction accuracy (Accuracy: 97.22\%, AUC: 0.9850). \\ 
\hline
\cite{wang_multiview_2021} & Adaptive heterogeneous multiview graph learning & Improved forecasting accuracy (Accuracy: 87.44\%). \\ 
\hline
\cite{jianying2021evaluation} & Optimized Backpropagation Neural Network (BPNN) and Particle Swarm Optimization-Backpropagation (PSO-BP) & Outperformed the single Backpropagation (BP) (MAE: 2.1022, $R^2$: 0.932). \\ 
\hline
\cite{chen_using_2021} & LSTM for forecasting & LSTM model suitable for risk prediction (MAE: 1.037 - 1.52). \\ 
\hline
\cite{zhu2016predicting} & RS-RAB & Best prediction performance (Accuracy: 86.74\%). \\ 
\hline
\cite{bassiouni_advanced_2023} & Various DL approaches & DL models improved accuracy by 14.46\% (Softmax: 76.32\%, SVM: 85.23\%). \\ 
\hline
\cite{zhu_forecasting_2019} & RS-MultiBoosting & superior performance of the RS-MultiBoosting ensemble method in forecasting \\ 
\hline
\cite{athaudage_modelling_2022} & RF regression & RF suitable for price prediction (MAE: 1.077, RMSE: 1.56, MAPE: 4.03\%). \\ 
\hline
\cite{liu_supply_2020} & AdaBoosted SVM (EN-AdaPSVM) & Outperforms other SVM models in noise data (Total False Rate: 11.76\% (EN-AdaPSVM), 13.41\% (AdaPSVM), 32.94\% (CSVM)). \\ 
\hline
\cite{zhu_comparison_2017} & Ensemble methods (RS-boosting and multi-boosting) & RS-boosting outperforms other methods (Average Accuracy: 85.41\% (RS-boosting), 84.08\% (multi-boosting), 74.80\% (boosting)). \\ 
\hline
\cite{fu_internet-based_2022} & BPNN, SVM, GA & BP-GA model outperforms BP and SVM (Accuracy: BP-GA 97.19\%, BP 82.33\%, SVM 95.31\%). \\ 
\hline
\cite{zhang_enterprise_2022} & Multimodal DL (DeepRisk) & DeepRisk outperforms baseline methods. \\ 
\hline
\cite{yang_identifying_2021} & Lasso-logistic model & Lasso-logistic model has best prediction accuracy (Prediction Accuracy: 96.5\%, Type II Error: 0.037). \\ 
\hline
\cite{zhang_credit_2022} & Fuzzy SVM (FSVM) & FSVM outperform traditional SVM and DT (Higher accuracy in training and test sets). \\ 
\hline
\cite{wang_research_2021} & Blockchain and fuzzy neural networks & Average Accuracy: 84.11\%, 84.66\%(Independent Variables Set Vs). \\ 
\hline
\cite{li_predicting_2022} & BPNN, LR & BPNN and LR provide accurate predictions (Prediction Accuracy: 92.8\%). \\ 
\hline
\cite{bodendorf_mixed_2023} & DL model with causal inference & High predictive performance and causal analysis. \\ 
\hline
\cite{liu_financial_2022} & Hybrid model (XGBoost–SMOTENC–RF) & Development of a hybrid model chain, incorporating XGBoost, SMOTENC, and RF models (overall accuracy rate 91.67\% and the average percentage error is 6.39\%). \\ 
\hline
\cite{rao_risk_2022} & LR, DT, integrated Logistic-RS & Integrated RS improves prediction performance (Accuracy of Logistic: 0.831, Accuracy of Logistic-RS: 0.871, Accuracy of DT: 0.772, Accuracy of DT-RS: 0.795). \\ 
\hline
\cite{handfield_assessing_2020} & Newsfeed analysis & Predicted factory risks for a five-year planning horizon. \\ 
\hline
\cite{pan_dynamics_2023} & DL BPNN & BPNN has high accuracy and low relative error (The maximum relative error calculated by the AHP is 57.41\%, and the minimum is 5.88\%, while the maximum relative error obtained by the BPNN is only 0.00210526\%). \\ 
\hline
\cite{wong_artificial_2022} & PLS-SEM, ANN & AI-based models (AIRM) are significant determinants of RP and SCA. \\ 
\hline
\cite{wang_e-commerce_2022} & Interval Type 2 Fuzzy Neural Network (IT2FNN) & IT2FNN punishes periodic deception, high accuracy (Accuracy 89\%). \\ 
\hline
\cite{nezamoddini_risk-based_2020} & GA with NN & Improved profits, lower inventory levels. \\ 
\hline
\cite{xia_predicting_2023} & SVM, RF, MLP, LR & RF showed best accuracy and low type I error. \\ 
\hline
\cite{zhao_credit_2022} & SVM and BPNN & Effective risk measurement, low relative error. \\ 
\hline
\cite{li_prediction_2022} & PCA-GA-SVM & PCA-GA-SVM outperforms SVM and GA-SVM, high accuracy. \\ 
\hline
\cite{duan_study_2021} & BPNN optimized by GA & Better fitting effect, higher prediction precision, and higher convergence speed. \\ 
\hline
\cite{zhang_firefly_2021} & SVM optimized by firefly algorithm (FA-SVM) & Improved classification efficiency and reduced error rates. \\ 
\hline
\cite{luo_construction_2022} & SVM, Particle Swarm Optimization, AdaBoost & Improved classification effect, highest accuracy of 96.13\%. \\ 
\hline
\cite{dang_evaluating_2022} & DL technology (BPNN) and blockchain & Effective application of DL and blockchain. \\ 
\hline
\cite{hosseini_bayesian_2016} & Bayesian network (BN) & BN allows disruption and improvement analysis. \\ 
\hline
\cite{gao_analysis_2022} & Autoregressive integrated (AR), Mixture Density Networks (MDN) model, Autoregressive Integrated Moving Average (ARIMA) model, Multilayer Perceptron-Long short term memory (MLP-LSTM) model, Particle Swarm Optimization (PSO) algorithm, and the improved PSO (IPSO) algorithm & AR-MDN outperforms other models, IPSO has strong performance. \\ 
\hline
\cite{wei_machine_2022} & ML-based linear regression algorithm (ML-LRA) & ML-LRA reduces supplier credit risk. \\ 
\hline
\cite{zheng_federated_2023} & Federated learning with dimensionality reduction (PCA) & Collective risk prediction benefits organizations with small datasets (Maximum Difference: 0.0670 (FL vs. Local Learning), 0.0634 (FL vs. Centralized Learning). \\ 
\hline
\cite{deiva_ganesh_supply_2022} & Text-mining & Use of social media data for contemporary risk understanding. \\ 
\hline
\cite{zhang_using_2023} & Time-decayed long short-term memory (TD-LSTM), Deep Neural Network & Improved predictability through TD-LSTM interpolation. \\ 
\hline
\cite{baryannis_predicting_2019} & SVM and tree learning & Framework offer good performance with interpretable and black-box methods (SVM Accuracy: 0.943, DT Accuracy: 0.950, Restricted DT (RDT) Accuracy: 0.916). \\ 
\hline
\cite{han_optimization_2021} & BPNN model & BPNN effectively predicted high to extremely high risks in a food production company's SC. \\ 
\hline
\cite{belhadi_ensemble_2021} & Hybrid EML approach: Gama Test (GT), Rotation Forest algorithm (RotF), Loogit Boosting algorithm (LB) & Good performance of RotF-LB model (Best results for RotF-LB: Accuracy 91.72, Precision 92.96, Recall 95.53, F-measure 94.23, AUC 92.89). \\ 
\hline
\cite{wu2022credit} & BPNN and GA-BPNN Credit Risk Model & Comparing iterations and risk prediction accuracy (GA-BPNN risk prediction accuracy above 0.92). \\ 
\hline
\cite{lei_research_2023} & Binary image difference common area (BCAoID), SVM-SMA & Highlight the effectiveness of the CGOA-SVM-SMA algorithm (Precision: 85.38\%, F-score: 63\%, TNR: 72\%). \\ 
\hline
\cite{yin_convolutional_2022} & Convolutional neural network model (CNN) & Optimal accuracy at 200 iterations (Comprehensive accuracy: 94.7\%). \\ 
\hline
\cite{janjua_fuzzy_2023} & Bi-LSTM CRF model & Comparing Bi-LSTM CRF with Baseline CRF (the Bi-LSTM CRF model achieved an overall F1-Score of 85\%, the Baseline CRF model achieved an overall F1-Score of 80\%). \\ 
\hline
\cite{yao_novel_2022} & Hybrid model combining FS-MRI, SVME-AIR, and SMV & AUC and KS scores for different parameters (AUC score of 0.8772 when the parameter ``num\_AIR" was set to 10, KS score of 0.6363 when the parameter ``num\_AIR" was set to 4). \\ 
\hline
\cite{zhang_credit_2015} & SVM and BPNN & SVM model is superior to the BPNN model (BPNN Accuracy 66.67\%, SVM model Accuracy 76.67\%). \\ 
\hline
\cite{aboutorab_reinforcement_2022} & RL-PRI approach & Time efficiency of RL-PRI vs. manual process (Tarif Dispute Accuracy: 0.941463, Recession Accuracy: 0.907937, Sea Level Rise Accuracy: 0.964286, Flood Accuracy: 0.856322). \\
\hline
\cite{wu_text-based_2024} & NLP techniques, word embedding models, bag-of-words approach & Developed SCRisk measure, significantly correlated with operational buffers and stock return volatility. \\
\hline
\cite{xu_enterprise_2024} & Large language models (e.g., GPT-4, BERT) & Improved decision support, optimized inventory management, reduced transportation costs, risk prediction. \\
\hline
\cite{chen2024using} & GA-BPNN, LR & High accuracy in predicting MSME default and credit ratings (Accuracy: 92\% for default prediction, 86\% for credit rating). \\
\hline
\cite{zhang2025domain} & Domain-Adaptation-based Multistage Ensemble Learning (DAMEL) & DEMEL outperformed traditional models in SME credit risk evaluation, addressing data scarcity, distribution discrepancies, and class imbalance (AUC: 0.7652). \\
\hline
\cite{zhou2025enhancing} & XGBoost, RF, CNN, Least Squares SVM (LSSVM) & XGBoost outperformed all models with accuracy: 91.9\%, precision: 92.8\%, recall: 94.7\%, F1-score: 93.7\%. SHAP analysis showed asset-liability ratio, quick ratio, and ESG ratings as key risk indicators. \\
\hline
\cite{mojdehi2025novel} & BallMapper + Graph Neural Network (BM-GNN) & BM-GNN effectively captures financial indicators and network-based features, improving SME credit risk assessment in SC Finance. \\
\hline
\end{longtable}

\normalsize

\section{Content Analysis of AI Techniques in SCRA}
\label{sec:AI Techniques}
In recent years, the application of ML methods in SCRA has attracted substantial attention due to its potential to improve decision-making processes and mitigate operational risks. This review paper explores various ML techniques applied in SCRA, highlighting their methodologies, key findings, and contributions.

\subsection{Used ML Models and Key Findings}
Table \ref{tab:ml_methods} summarizes various research methods or models used in risk prediction and analysis, along with the key findings and associated metrics from different research papers. Each row in the table represents a research paper's first author, the research method or model used, and the specific findings and metrics achieved. These methods and findings are crucial in understanding risk prediction in various domains.

\subsection{AI Models Used in SCRA}
\begin{figure*}[!ht]
    \centering
    \includegraphics[width=1\textwidth]{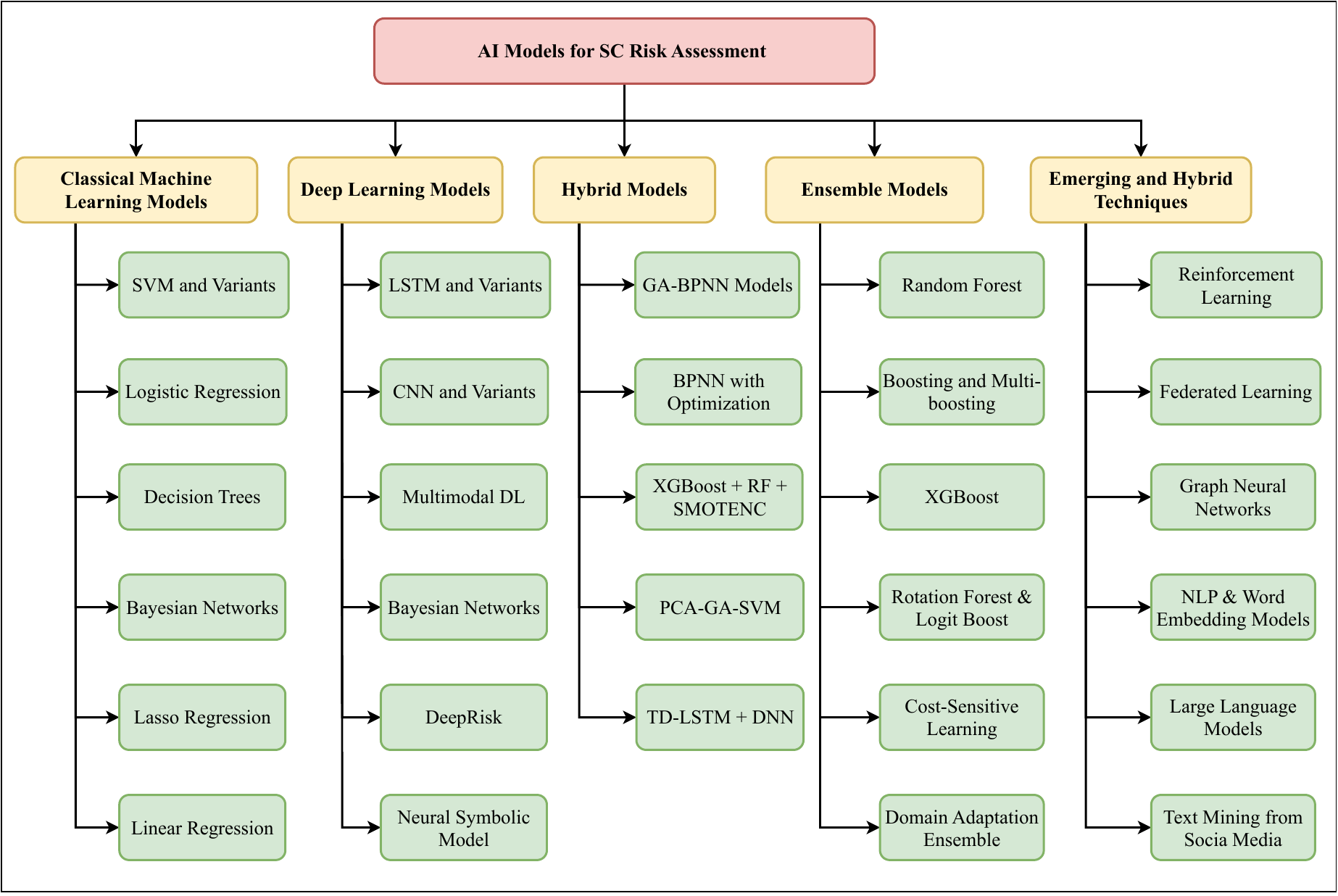}
    \caption{Classification of AI models in SCRA based on methodology, complexity, and common application scenarios.}
    \label{fig:classification}
\end{figure*}

A wide range of ML and AI techniques have been applied to SC risk prediction, reflecting the increasingly data-driven nature of modern risk management. These techniques can be broadly classified into five clusters: Classical ML (CML) models, DL models, Hybrid CML+DL approaches, Ensemble methods, and Emerging Techniques such as Federated Learning, Reinforcement Learning, and Large Language Models (LLMs). Each cluster offers distinct advantages, limitations, and application contexts, with the literature providing diverse empirical evidence supporting their use. Each category is defined based on its distinct methodological characteristics, level of complexity, application scope, and suitability for specific SC risk scenarios, as shown in Figure~\ref{fig:classification}. 

A detailed overview of each AI model used in SCRA, including its key characteristics, advantages, disadvantages, application cases, and suitable scenarios, is given below:

\subsubsection{CML Models}
CML models remain popular in SC risk prediction due to their simplicity, interpretability, and computational efficiency. Models such as Support Vector Machines (SVM), Logistic Regression (LR), Decision Trees (DT), and Bayesian Networks (BN) are particularly effective when working with structured, tabular data where relationships between variables are relatively well understood. For example, SVM and its variants have been used successfully in several studies for credit risk assessment, supplier default prediction, and operational risk evaluation \citep{zhu2016predicting, zhang_credit_2015, zhao_credit_2022, li_prediction_2022, zhang_firefly_2021, liu_supply_2020, zhang_credit_2022, fu_internet-based_2022, liu_financial_2022, xia_predicting_2023, lei_research_2023, gao_analysis_2022, yao_novel_2022}. These models, often improved through feature selection or optimization algorithms, achieve high classification accuracy, particularly when combined with techniques such as genetic algorithms (GA) and principal component analysis (PCA). Logistic regression remains a favored baseline due to its transparency and ease of interpretation, applied in risk assessment settings where explainability is required, such as in \cite{li_predicting_2022, rao_risk_2022, chen2024using}. Decision Trees (DTs), employed in studies such as \cite{baryannis_predicting_2019, rao_risk_2022}, provide interpretable decision rules but often overfit unless combined into ensembles. Meanwhile, Bayesian Networks (BN) offer a probabilistic, causal perspective on risk propagation, particularly in disruption analysis settings, as demonstrated by \cite{hosseini_bayesian_2016}. Lasso regression models, which excel in feature sparsity, were applied in \cite{yang_identifying_2021} to isolate key predictors of SC risk. ML-based linear regression algorithm (ML-LRA) proposed by \cite{wei_machine_2022} reduces supplier credit risk. Though highly interpretable, these models often struggle with capturing non-linear relationships and require significant domain expertise to manually engineer relevant features, limiting their scalability in high-dimensional or unstructured data environments. CML models offer simple implementation and training, interpretability that supports regulatory compliance, and are suitable for smaller datasets. However, these models have a limited ability to capture complex, non-linear patterns and may require extensive feature engineering. They are commonly used for early-stage risk identification, for interpretable models in regulatory audits, and in situations with limited historical data.

\subsubsection{DL Models}
In contrast, DL models have demonstrated superior performance in capturing complex, non-linear patterns in SC risk prediction, particularly when large, high-dimensional, or unstructured data. LSTM networks, known for their strength in sequence modeling, have been widely applied for time series forecasting in SC finance and disruption prediction \citep{chen_using_2021, zhang_using_2023, janjua_fuzzy_2023, gao_analysis_2022}. CNN has also been adopted, particularly when analyzing spatially structured data or feature maps, as seen in \cite{yin_convolutional_2022, zhou2025enhancing, zhang_using_2023}. Multimodal DL \citep{zhang_enterprise_2022,bassiouni_advanced_2023,bodendorf_mixed_2023,pan_dynamics_2023} models have emerged to integrate diverse data types, such as financial indicators, operational metrics, and external news sources, to improve predictive performance, with DeepRisk showing strong results in \cite{zhang_enterprise_2022}. More recently, tensorization-based neural symbolic models \citep{katsaliaki_supply_2022} have been proposed, blending symbolic reasoning with DL for relational risk prediction. While DL models demonstrate state-of-the-art accuracy, their application in SCs is hampered by high computational costs, data-hungry training processes, and their opaque, black-box nature, which hinders regulatory acceptance in some industries. DL models provide superior performance with complex, high-dimensional data, effective handling of temporal and spatial dependencies, and the potential to combine multimodal data. However, they come with high computational costs, poor interpretability without dedicated explainability tools, and the need for large labeled datasets. They are well-suited for high-frequency sensor data (SC), multimodal data integration (text, time series, images), and long-term risk forecasting.

\subsubsection{Hybrid Models (CML + DL)}
Hybrid approaches combining classical ML and DL techniques aim to strike a balance between interpretability and predictive accuracy, particularly in scenarios where data characteristics vary across dimensions or stages of the risk prediction process. Several studies combined genetic algorithms (GA) with Backpropagation Neural Networks (BPNN), resulting in improved convergence rates and prediction precision \citep{wu2022credit, fu_internet-based_2022, nezamoddini_risk-based_2020, duan_study_2021, chen2024using}. In \cite{jianying2021evaluation}, PSO was also used to optimize BPNN parameters \citep{dang_evaluating_2022,han_optimization_2021}, outperforming single models. Hybrid model chains, such as the XGBoost-SMOTENC-RF pipeline proposed in \cite{liu_financial_2022}, improve performance in class-imbalance scenarios common in SC risk datasets. A particularly novel hybrid, TD-LSTM combined with deep neural networks (DNN), demonstrated strong forecasting accuracy in SC financial data in \cite{zhang_using_2023}. By combining domain-specific feature engineering with data-driven learning, these hybrid approaches address non-linearity while preserving interpretability, though they introduce higher complexity and tuning costs. Hybrid models offer a balance between interpretability and predictive power, flexibility to handle different data types and distributions, and often greater robustness to overfitting. However, they come with increased model complexity, longer training times, and the risk of suboptimal tuning in heterogeneous frameworks. They are useful for medium-sized datasets with mixed data types, cases requiring both prediction accuracy and interpretability, and data with substantial noise and missing values.

\subsubsection{Ensemble Models (Bagging, Boosting, Stacking)}
Ensemble methods, particularly RF \citep{athaudage_modelling_2022,wang_forecasting_2022,wang_multiview_2021,xia_predicting_2023,liu_financial_2022,zhu_forecasting_2019}, boosting algorithms (XGBoost, AdaBoost) \citep{liu_financial_2022,zhou2025enhancing,zhang2025domain}, and multi-boosting approaches consistently outperform individual SC risk prediction models by aggregating predictions from multiple weak learners. Studies such as \cite{zhu_forecasting_2019, zhu_comparison_2017, liu_supply_2020, belhadi_ensemble_2021} demonstrated that multi-boosting ensembles effectively handle data noise and feature redundancy, improving accuracy in risk classification tasks. Cost-sensitive Random Forest (CSL-RF) models further improve robustness under unbalanced risk categories, as shown in \cite{wang_forecasting_2022}. Novel approaches, such as the Domain-Adaptation-based Multistage Ensemble Learning (DAMEL) framework proposed in \cite{zhang2025domain}, directly address data scarcity, distribution shifts, and class imbalance, which are particularly relevant for SME credit risk prediction. While ensembles excel in predictive accuracy, their major trade-off is reduced interpretability, especially as ensemble size increases.
Ensemble models provide improved predictive performance, greater robustness to overfitting, and the ability to combine the strengths of multiple algorithms. However, they are computationally intensive, poorly interpretable, and difficult to tune optimally. They are ideal for large, noisy datasets, situations where prediction accuracy is paramount, and multi-source data environments.

\subsubsection{Emerging and Hybrid Techniques}
Several cutting-edge techniques have emerged, demonstrating strong potential for real-time risk monitoring and collaborative prediction across SC partners. Federated learning (FL) approaches, such as those in \cite{zheng_federated_2023}, enable decentralized risk modeling across firms without exposing sensitive data, a crucial advantage in privacy-sensitive SC networks. Reinforcement Learning (RL), as employed in \cite{aboutorab_reinforcement_2022}, optimizes adaptive intervention strategies, balancing risk-mitigation costs and performance outcomes. Graph Neural Networks (GNNs), such as the BM-GNN framework introduced in \cite{mojdehi2025novel}, excel at capturing relational dependencies among suppliers, thereby enriching risk-propagation modeling. Lastly, natural language processing (NLP) and LLMs, including GPT-4 and BERT, have been applied to text-based risk identification by extracting risk signals from news, social media, and operational reports \citep{wu_text-based_2024, xu_enterprise_2024, deiva_ganesh_supply_2022}. These emerging techniques enable collaborative modeling across multiple organizations, preserve data privacy, and offer hybrid symbolic-ML interpretability. However, they require careful coordination among participants, are vulnerable to communication bottlenecks, and may suffer from drift across sites during model updates. They are well-suited for collaborative supplier monitoring, risk prediction across supply networks, and situations where data privacy regulations restrict data sharing.

\section{Practical and Managerial Implications}
\label{sec:Managerial Implications}

The findings of this study offer important insights for practitioners aiming to improve SCRM using AI and ML. In today’s environment, supply chains are becoming more complex, highly interconnected, and increasingly exposed to disruptions. As a result, organizations need to move beyond traditional reactive approaches and adopt more proactive and predictive risk management strategies based on data-driven insights.

\subsection{Strategic Implications}  
At the strategic level, this study shows that AI and ML are no longer just supporting tools but have become essential for modern supply chain risk assessment (SCRA). Advanced models such as Random Forest (RF), XGBoost, and hybrid approaches have shown strong capability in predicting risks more accurately, helping organizations take early and informed actions \citep{zhou2025enhancing}. However, these benefits can only be achieved when organizations have access to reliable and well-integrated data. This means that firms must invest in proper digital infrastructure and effective data management systems. After COVID-19, the need for flexibility and adaptability has become more important than ever \citep{sheng_covid-19_2021}. Organizations should integrate AI-based risk assessment into their overall resilience strategies, focusing on contingency planning and scenario analysis. This reflects a shift from fixed and static risk management approaches to more dynamic systems that can continuously adjust to changing conditions. Companies that successfully align AI with their strategic planning are more likely to build stronger resilience and maintain long-term competitiveness. In addition, combining blockchain technology with ML models can further improve transparency and trust within supply chains \citep{dang_evaluating_2022}. Blockchain helps ensure secure and reliable data sharing, while ML supports predictive analysis. Together, they create a more robust and collaborative risk management system.

\subsection{Tactical Implications}  
At the tactical level, choosing the right AI models is very important for effective risk assessment. The findings indicate that ensemble methods like RF and XGBoost perform well in handling complex data and improving prediction accuracy \citep{zhou2025enhancing}. Time-series models such as ARIMA and LSTM are also useful for forecasting demand changes and identifying potential disruptions based on past trends \citep{gao_analysis_2022}. Hybrid models, which combine different ML techniques, provide a more complete view of supply chain risks by capturing both simple and complex relationships \citep{mojdehi2025novel}. These approaches help managers design more flexible and reliable risk assessment systems. However, in practice, managers should not focus only on model accuracy. Other factors such as ease of understanding, scalability, and compatibility with existing systems are equally important. Decision-makers need to clearly understand the results produced by these models in order to take effective actions. Therefore, explainable AI plays a key role in making advanced models more usable in real-world decision-making.

\subsection{Operational Implications}  
At the operational level, AI-based SCRA allows organizations to monitor risks in real time and detect potential disruptions at an early stage. The use of real-time and alternative data sources, such as social media platforms (e.g., Twitter), helps in identifying emerging risks quickly \citep{deiva_ganesh_supply_2022}. This enables organizations to act early instead of reacting after the disruption occurs. Risk predictions generated by AI should be integrated into operational systems, such as dashboards and decision-support tools, to support timely actions. These may include adjusting inventory levels, selecting alternative suppliers, or modifying production plans. Regular evaluation and comparison of different ML models are also necessary to ensure that the most suitable methods are being used. From a practical viewpoint, AI insights should be embedded into daily operations rather than used separately. Organizations that successfully integrate real-time risk information into their workflows can respond faster and reduce the impact of disruptions. SMEs require special consideration due to their limited resources and unique operational constraints. In such contexts, collaborative and privacy-preserving approaches to risk assessment are particularly important, as they enable organizations to use shared data insights without compromising sensitive information \citep{dang_evaluating_2022,wu_credit_2022,yang_supply_2023}. At the same time, organizations should invest in training their employees in data analytics and AI to ensure effective use of these technologies. It is also important to follow data privacy and regulatory requirements to avoid potential legal issues. Overall, the study suggests that adopting AI for supply chain risk assessment is not only a technological decision but also an organizational one. Success depends on proper planning, alignment with business strategy, and continuous learning. Managers should treat AI as an integral part of their supply chain risk management system rather than as an isolated tool.

\section{Challenges and Limitations}
\label{sec:challenges}
Despite the growing adoption of AI and ML in SCRA, several challenges and limitations remain that affect both practical implementation and future research development. These challenges are not only technical in nature but also organizational and operational, making their resolution more complex.

\subsection{Data Quality and Availability}  
One of the most critical challenges in AI-driven SCRA is the availability of high-quality and reliable data. In real-world supply chains, data is often scattered across different systems, stored in inconsistent formats, and affected by missing or noisy values. Such issues directly reduce the effectiveness of ML models and can lead to inaccurate risk predictions \citep{araz_data_2020,baryannis_supply_2019,ni_systematic_2020}. From a practical perspective, this problem becomes more serious because supply chain decisions are highly time-sensitive. Decisions based on poor-quality data may lead to incorrect demand forecasts, supplier misselection, or delayed responses to disruptions. Moreover, many organizations still lack proper data integration across suppliers, logistics partners, and internal systems. Therefore, improving data governance, standardization, and real-time data integration is essential for making AI-based risk assessment truly effective.

\subsection{Model Interpretability and Transparency} 
Another important limitation is the lack of interpretability in many advanced AI models. Techniques such as deep learning and ensemble models often function as black boxes, making it difficult for managers to understand how risk predictions are generated. This creates a major barrier in supply chain decision-making, where accountability and justification are required. Although prior studies emphasize the importance of explainable AI in supply chain risk management, its practical adoption remains limited \citep{nimmy_explainability_2022,olan_enabling_nodate}. In real organizational settings, managers are less likely to rely on models they do not understand, even if those models are highly accurate. This highlights a key trade-off between model accuracy and usability. To ensure wider adoption, future research should focus on developing models that are both accurate and interpretable, allowing decision-makers to confidently act on AI-generated insights.

\subsection{Data Imbalance and Rare Events}  
Supply chain disruptions are typically rare but have a high impact, which leads to highly imbalanced datasets. Most ML models are trained on historical data where normal operations dominate, causing them to perform poorly in predicting rare disruption events \citep{shi_machine_2022,liu_supply_2020}. In practice, this limitation is critical because rare events such as supplier failures, pandemics, or geopolitical disruptions often cause the most severe consequences. If models fail to capture these events, organizations may remain unprepared for major risks. This indicates the need for more advanced techniques, such as resampling methods and anomaly detection models, that can better identify low-frequency but high-impact risks.

\subsection{Integration with Existing Systems}  
The integration of AI models into existing supply chain systems is another major challenge. Many organizations still rely on legacy systems that are not designed to support advanced analytics or real-time data processing. This creates difficulties in deploying ML models effectively within operational workflows. Research shows that successful digital transformation in supply chains requires alignment between technology, processes, and organizational structure \citep{schroeder_systematic_2021,wang_education_2024}. In practice, even well-developed models may fail if they cannot be integrated into decision-support systems or if employees are not trained to use them. Therefore, organizations need to focus not only on model development but also on system compatibility, process redesign, and change management.

\subsection{Computational Complexity and Resource Constraints}  
Advanced AI models, especially deep learning and hybrid approaches, require significant computational power, large datasets, and technical expertise. This creates a barrier for many organizations, particularly small and medium-sized enterprises (SMEs), which often lack such resources \citep{janiesch_machine_2021,nayal_are_2021}. From a managerial perspective, this creates a practical trade-off between model sophistication and implementation feasibility. While complex models may provide higher accuracy, they may not always be suitable for real-world use due to cost and infrastructure limitations. This highlights the need for scalable, cost-effective, and user-friendly AI solutions that can be adopted across different types of organizations.

\subsection{Generalizability and Context-Specific Limitations}  
Another limitation is the lack of generalizability of many AI models. Models developed using data from a specific industry or region may not perform well in different supply chain contexts due to variations in operational structures, risk factors, and market conditions \citep{xu_disruption_2020,yang_supply_2023}. In real-world applications, this means that organizations cannot directly adopt existing models without proper validation and adaptation. Supply chain risks are highly context-dependent, and a model that works well in one scenario may fail in another. Therefore, future research should focus on developing more flexible and adaptable models, as well as conducting cross-industry validation studies.

\subsection{Ethical and Data Privacy Concerns}  
The use of AI in supply chain risk assessment also raises important ethical and data privacy concerns. Supply chains often involve multiple stakeholders, and sharing sensitive operational or financial data can create risks related to confidentiality and security. Existing studies emphasize the importance of secure and privacy-preserving approaches in AI-based supply chain systems \citep{dang_evaluating_2022,wu_credit_2022}. In practice, failure to address these concerns can lead to regulatory issues, loss of trust, and reduced collaboration among supply chain partners. Therefore, organizations must ensure compliance with data protection regulations and adopt secure data-sharing mechanisms to support collaborative risk management. While AI-driven SCRA offers strong potential for improving risk prediction and decision-making, these challenges highlight that its successful implementation requires more than just technological capability. It requires organizational readiness, proper data management, and continuous adaptation. Addressing these limitations will be essential for fully realizing the benefits of AI in supply chain risk management.

\section{Future Research Directions}
\label{sec:Future Research Directions}
In light of the challenges and limitations, future research directions in AI-based SCRA are multifaceted and have the potential to advance the state of the art. The development of AI models that deliver accurate predictions and offer comprehensive transparency and interoperability should be emphasized. Techniques like Local Interpretable Model-Agnostic Explanations (LIME) and SHapley Additive exPlanations (SHAP) serve as pivotal approaches to improve the interpretability of AI models \citep{olan_enabling_nodate}. By employing these methods, the goal is to demystify AI models' decision-making processes, enabling stakeholders to understand the rationale behind specific predictions or decisions. This transparency improves trust and facilitates more effective integration of AI-driven insights into SC risk management strategies. 

Combining multiple AI techniques, such as ML and expert systems, can improve the robustness of SCRA. Future research should explore hybrid AI models that leverage the strengths of different approaches to improve prediction accuracy and reliability \citep{mojdehi2025novel}. Addressing data quality challenges involves multifaceted strategies. Utilizing larger datasets for analysis, as emphasized by \citep{wang_multiview_2021}, improves the robustness of risk assessment, enabling a more comprehensive evaluation of risk factors and improving predictive accuracy. Additionally, initiatives should aim to democratize access to standardized SC data repositories. Creating and maintaining such repositories enables stakeholders across the SC spectrum to access reliable and consistent data. Increased accessibility fosters better-informed decision-making and promotes the use of standardized datasets for improved AI model training and validation. Investigating AI models capable of real-time risk assessment will be pivotal. These models should adapt to emerging risks and provide timely alerts or recommendations for risk mitigation.

Research on creating integration frameworks for AI-based SCRA is essential \citep{shahzadi_ai_2024}. These frameworks should facilitate easy integration into existing SC management systems, ensuring a smooth transition to AI-driven risk assessment. Expanding the application of AI in risk assessment to various domains within the SC, including logistics, procurement, and production, offers opportunities for comprehensive risk management. Investigate AI models capable of adapting to pandemic-induced disruptions within SCs. Focus on understanding the unique dynamics and challenges posed by COVID-19, aiming to develop AI-driven strategies that enable resilience and agility in such crisis scenarios. As AI plays an increasingly significant role in decision-making, researchers should also address ethical and legal considerations. This includes issues related to data privacy, bias, and fairness in AI models. By addressing these challenges and pursuing these research directions, AI-based SCRA can make significant strides toward more effective and reliable risk management in an ever-evolving global SC landscape.

\section{Conclusions}
\label{sec:conclusion}
This extensive review explored the SCRA domain and examined the complex challenges of integrating AI, especially ML. Our thorough analysis encompassed a large pool of 1,903 SCRA papers, ultimately narrowing our focus to a refined subset of 54. This endeavor has yielded significant insights and key findings, making substantial contributions to the SCRA literature.

Our primary contributions can be summarized by addressing the four key research questions:
\begin{itemize}
\item[RQ-1:] Our review thoroughly analyzes the existing state of research concerning the application of AI and ML techniques in SCRM. We have critically assessed a substantial body of literature to extract meaningful insights, specifically from 54 selected articles.
\item[RQ-2:] Through our analysis, we have identified the AI and ML techniques most commonly utilized in the domain of SCRA. This clarifies the methodologies employed and underscores the trends shaping the field.
\item[RQ-3:]  Our review highlights the key findings and trends prevalent in the literature, shedding light on the progress, developments, and major areas of focus within AI and ML in SCRM.
\item[RQ-4:] We have identified critical research gaps and proposed future directions to guide further explorations in this field, ensuring the continued evolution of SCRA methods.
\end{itemize}

Despite these contributions, our research has some limitations. Our data primarily originated from Google Scholar and WoS rather than Scopus, potentially introducing biases. Our focus on papers published between 2015 and 2025 presents a limitation, as it may omit earlier relevant research that could provide valuable historical context. Additionally, our restriction to English-language papers may have omitted valuable non-English research.

Looking forward, several promising avenues for future research emerge in the domain of SCRA using AI. These directions encompass operationalizing interconnectedness, transformability, and sharing within SCRA frameworks; investigating evolving ICT roles in prediction and response; revisiting the cost implications of resilience improvement; and exploring emerging technologies such as blockchain. These endeavors are expected to improve the effectiveness of risk assessment in the post-COVID era. Reviewers should adopt a comprehensive approach to gain deeper insights into this field by expanding their search to diverse databases, including non-English and grey literature sources, using snowballing techniques, collaborating with experts from related disciplines, leveraging ML tools for data analysis, and staying up to date with the latest research. By combining various search methods and expert opinions, this multidisciplinary strategy is essential for uncovering valuable insights and emerging trends in this dynamic and critical field.

\section*{Declarations}

\subsection*{Ethical Approval}
This study is a comprehensive review of existing literature and did not involve any new studies of human or animal subjects performed by any authors. Therefore, it did not require any ethical approval or informed consent.

\subsection*{Funding}
No funding was received to conduct this study.

\subsection*{Competing Interests}
The authors have no financial or proprietary interests in any material discussed in this article.

\subsection*{Availability of Data and Material}
Data sharing does not apply to this article as no datasets were generated or analyzed during the current study. This study is a comprehensive review of existing literature. No new human data were directly used or accessed by the authors of this review.

\subsection*{Authors' Contributions}
\textbf{Md Abrar Jahin}: Conceptualization, Data curation, Formal analysis, Investigation, Methodology, Software, Writing - original draft, Visualization.
\textbf{Saleh Akram Naife}: Data curation, Formal analysis, Investigation, Methodology, Writing - original draft.
\textbf{Anik Kumar Saha}: Data curation, Methodology, Writing - original draft.
\textbf{M. F. Mridha}: Validation, Writing - review \& editing.

\subsection*{Acknowledgements}
The authors would like to thank the University of Southern California (USC) for providing the unlimited remote access to a wide range of closed-access journals and institutional support that made this research possible.

%%%% don't change the commands below %%%%%%%
%\twocolumn
\bibliographystyle{apalike}
\bibliography{main}

% \linespread{1}
% \onecolumn 
% \appendix

\end{document}